\documentclass[14pt]{extarticle}
\usepackage[bookmarks=true]{hyperref}
\usepackage[title]{appendix}
\hypersetup{
  pdftitle={Both eyes open: Vigilant Incentives help Regulatory Markets improve AI Safety},
  pdfsubject={cs.CY, cs.AI},
  pdfauthor={Paolo Bova, Alessandro Di Stefano, The-Anh Han},
  colorlinks=true,       % false: boxed links; true: colored links
  linkcolor=blue,       % color of internal links
  citecolor=blue,       % color of links to bibliography
  filecolor=magenta,        % color of file links
  urlcolor=cyan,        % color of external links
  linktoc=page,            % only page is linked
  pdfpagemode=FullScreen,
}

\usepackage[
backend=biber,
style=apa,
sorting=nyt,
minsortnames=1,
maxsortnames=2,
date=year,
labeldate=year,
doi=true
]{biblatex}
\usepackage{tikz} % For MC
\usepackage{amsmath} % For MC Math definitions
\usepackage{IEEEtrantools} % For MC
\usepackage{enumitem} % For MC
\usepackage[capitalise,nameinlink,noabbrev]{cleveref}
\creflabelformat{equation}{#2\textup{#1}#3}
\usepackage[all]{nowidow}

\usepackage{arxiv} % arXiv template

\usepackage[T1]{fontenc}    % use 8-bit T1 fonts

\usepackage{authblk}

\graphicspath{{./images/}}
\usepackage{caption}
\title{Both eyes open: Vigilant Incentives help Regulatory Markets improve AI Safety}

\DeclareRobustCommand{\orcidicon}{
	\begin{tikzpicture}
	\draw[lime, fill=lime] (0,0) 
	circle [radius=0.16] 
	node[white] {{\fontfamily{qag}\selectfont \tiny ID}};
	\draw[white, fill=white] (-0.0625,0.095) 
	circle [radius=0.007];
	\end{tikzpicture}
	\hspace{-2mm}
}
\foreach \x in {A, ..., Z}{\expandafter\xdef\csname orcid\x\endcsname{\noexpand\href{https://orcid.org/\csname orcidauthor\x\endcsname}
			{\noexpand\orcidicon}}
}
\author[1\thanks{\tt{paolobova@protonmail.com}}]{Paolo Bova}
\author[1\thanks{\tt{A.DiStefano@tees.ac.uk}}]{Alessandro Di Stefano\orcidB{}}
\author[1\thanks{\tt{T.Han@tees.ac.uk}}]{The-Anh Han\orcidC{}}

\affil[1]{Teesside University \href{https://research.tees.ac.uk/}{https://research.tees.ac.uk/}}

\renewcommand{\shorttitle}{Vigilant Incentives help Regulatory Markets}

\usepackage{graphicx}
\usepackage{subcaption}
\usepackage[most]{tcolorbox}
\usepackage{booktabs}  % professional-quality tables
\usepackage{tabularx}
\usepackage{esvect}  % better vector notation

\begin{document}

\maketitle

\begin{abstract}

In the context of rapid discoveries by leaders in AI, governments must consider how to design regulation that matches the increasing pace of new AI capabilities. Regulatory Markets for AI is a proposal designed with adaptability in mind. It involves governments setting outcome-based targets for AI companies to achieve, which they can show by purchasing services from a market of private regulators. We use an evolutionary game theory model to explore the role governments can play in building a Regulatory Market for AI systems that deters reckless behaviour. We warn that it is alarmingly easy to stumble on incentives which would prevent Regulatory Markets from achieving this goal. These “Bounty Incentives” only reward private regulators for catching unsafe behaviour. We argue that AI companies will likely learn to tailor their behaviour to how much effort regulators invest, discouraging regulators from innovating. Instead, we recommend that governments always reward regulators, except when they find that those regulators failed to detect unsafe behaviour that they should have. These “Vigilant Incentives” could encourage private regulators to find innovative ways to evaluate cutting-edge AI systems. 

% We also use our model to visualise how a Regulatory Market achieves a comparable reduction in risk to an institutional regulator whilst minimising overregulation. We conclude that Regulatory Markets with well-designed incentives can complement institutional regulators.

\end{abstract}

\begin{tcolorbox}[colback=green!5!white,colframe=green!75!black , left=1em,right=2em, top=1em, bottom=2em]

\begin{center}
\large
\textbf{Highlights}
\end{center}

\tcblower
    
\begin{itemize}
    \bigskip
    \item We show that governments can incentivise a healthy Regulatory Market using what we call “Vigilant Incentives” \textemdash{} which always pay private regulators unless they fail to detect unsafe behaviour. On the other hand, “Bounty Incentives” \textemdash{} which pay only when they catch unsafe behaviour \textemdash{} destabilise Regulatory Markets.\\
    \item “Vigilant Incentives” are effective because AI companies are sensitive to how likely private regulators are to detect unsafe behaviour. This allows a Regulatory Market to act as a deterrent to neglecting AI Safety.\\
    \item We quantify how good regulators have to be at detecting unsafe AI systems to effectively deter reckless behaviour and highlight it as a crucial measure of the health of the Regulatory Market.\\
    \item We also assess the importance of the size of the incentives. To balance risk reduction and overregulation concerns, incentives should not be too generous, except in situations where large externalities suggest that we prioritise risk reduction.\\
    \item We visualise how Regulatory Markets are much better at balancing these tradeoffs under uncertainty than direct government regulation would be. However, Regulatory Markets also require a vigilant government regulator to assess the effects of the Regulatory Market. Regulatory Markets could hold promise in magnifying the impact of the government while minimising concerns of overregulation.
\end{itemize}

\end{tcolorbox}

\break

\section{Introduction}

A challenge facing us today is to find ways to govern the long-term development of new AI capabilities safely. AI researchers recognise that it will be more difficult to align the intentions and values of future goal-directed AI systems with those of the groups they serve \parencite{amodei2016concrete, leike2017ai, hern_andez_orallo2019surveying, krakovna2020specification}. Even if researchers can address these technical safety challenges, the deployment of powerful AI capabilities brings with it concerns of misuse, especially when we consider the dual use of many AI capabilities \parencite{brundage2018malicious, shevlane2019offense, zwetsloot2019thinking,}.

Governments around the world have begun to respond to these challenges. The NIST AI roadmap is an example of efforts by the United States to guide and promote industry self-regulation \parencite{tabassi2021artificial, barrett2022actionable}. The European Union’s AI Act, set to come into effect soon after years of refinement, pursues a more binding regulatory framework that some argue may influence future efforts elsewhere \parencite{siegmann2022brussels}.

For now, it is not clear whether these efforts will meaningfully reduce the risks from future AI capabilities. Meanwhile, as feedback submitted on the above projects suggests, AI companies are paying close attention to the future of international regulatory environments. There is time for new regulatory initiatives to take effect before AI companies commit to a development strategy. 

The field of AI governance has proposed many possible initiatives: The literature has iterated on several frameworks for auditing future AI systems: from Model Cards to System Cards, and audits that explicitly highlight the relevant effects of AI systems on their stakeholders \parencite{mitchell2019model, gursoy2022system, brown2021algorithm}. \textcite{cihon2021ai} explore AI certification schemes to enforce technical and ethical safety standards. \textcite{cihon2020should} have also explored the building of new technical standard-setting organisations. There are even discussions of novel voluntary agreements such as \citeauthor{o_keefe2020windfall}'s (\citeyear{o_keefe2020windfall}) Windfall Clause. In many cases, regulatory sandboxes have been suggested as a low commitment means to trial several of the new initiatives above. This only scratches the surface of the available menu of actions that governments and companies could consider \parencite{naud_e2020race, brundage2020trustworthy, cihon2021corporate, truby2022sandbox}.

Regulatory Markets present a relatively novel approach to regulation \parencite{clark2019regulatory}. Governments set targets and mandate that companies employ the services of private regulators to demonstrate compliance with those targets. These Regulatory Markets act as a complement and not a substitute for building government capacity to monitor the activities of AI companies and the capabilities of AI systems on the horizon \parencite{whittlestone2021why}.

Regulatory Markets have favourable qualities which seem appropriate for the uncertain and adaptable terrain of AI development. Private regulators must compete with each other to regulate AI companies. This competition may lead to innovations in methods to detect unsafe behaviour and better understand what safe development practises look like. Are these proposed benefits likely? Will any start-ups join the proposed Regulatory Market? 

Our paper makes three contributions, the first of which is to show under which conditions we can expect a Regulatory Market to be successful. We argue that well-chosen, appropriately funded regulators will participate in a Regulatory Market and can be incentivised to produce high-quality detection methods and standards. These regulators are not just effective in catching unsafe behaviour. They also act as an effective deterrent to unsafe behaviour.

Not all incentives will encourage high-quality regulators to join the Regulatory Market. Some incentives, for example those which encourage an adversarial relationship between AI companies and regulators, will actively harm the Regulatory Market. These incentives unfortunately have an appealing efficiency at first glance, since they focus on rewarding regulators for catching unsafe behaviour (we call such incentives “Bounty Incentives”). Incentives that instead appreciate the role of the Regulatory Market as a deterrent to unsafe behaviour fare much better (which we call “Vigilant Incentives”).

We arrived at this first conclusion by modelling the different incentives that would face both private regulators and the AI companies they regulate. This model builds on an existing model of the market for new AI capabilities, known as the DSAIR model \parencite{han2020regulate}. We extend this model to capture the detection and enforcement abilities of private regulators. Our work contributes to a growing number of publications that model competitive dynamics in AI markets \parencite{armstrong2016racing, askell2019role, han2020regulate, naud_e2020race, lacroix2022tragedy}. To capture the complex dynamics that may emerge as regulators and companies explore the strategy space, we turn to analytical and numerical methods from Evolutionary Game Theory \parencite{foster1990stochastic, fudenberg2006evolutionary,, wallace2015stochastic}. Evolutionary Game Theory has been used to study other incentive mechanisms, both for issues in AI Governance, and in Climate Change, another issue characterised by high uncertainty and multiple types of actors \parencite{han2020regulate, lacroix2022tragedy, encarna_cc_ao2016paradigm, santos2016evolutionary}.

As a second contribution, we discuss the trade-offs that one might consider when funding a Regulatory Market: as with other forms of regulation, deterring more unsafe behaviour often has the side effect that regulators are more likely to slow down companies in scenarios where the risks are low, an outcome we call “overregulation” in line with prior work \parencite{han2020regulate, han2021mediating, han2022voluntary}.

Here, we invoke the double-blind problem of the Collingridge Dilemma \parencite{worthington1982social}. Governments will likely know little about the capabilities and risks of new AI capabilities until those technologies become entrenched. At that point, it will probably be very difficult to influence who controls the market for AI. For this reason, measures such as a Regulatory Market must act under uncertainty.

In particular, if the risks are low enough and the speed advantage from neglecting safety norms is high enough, then Regulatory Markets will lead to overregulation. As the Collingridge dilemma implies, these are two parameters of our model that are highly uncertain, and there exists much disagreement about where different approaches to AI sit and whether it makes sense to see AI Safety as separate from AI Capabilities in the first place \parencite{cave2018ai, dafoe2018ai, burden2020exploring, vinuesa2020role}.

We find that we can reduce overregulation with little impact on risk through the careful design of government incentives and regulator activities. The nature of the externalities that AI systems pose can have a large influence on these designs.

For our final contribution, we compare Regulatory Markets to a government that directly regulates AI companies. Under uncertainty, we find that Regulatory Markets fare much better in balancing risk reduction and overregulation than the Government does. We also note that Vigilant Incentives require that governments maintain a strong capacity for monitoring the market for AI. 

The rest of the paper proceeds as follows: \cref{sec:model} outlines our model of Regulatory Markets, whereas \cref{sec:methods} describes the evolutionary game theory method we adopt. \cref{sec:results} discusses the above three results in more detail along with detailed figures. \cref{sec:discussion} discusses how policymakers might use the model as a tool for thinking about how to evaluate a future Regulatory Market. We conclude with a brief discussion of model limitations and possible future research directions.

\section{Model}\label[section]{sec:model}

This section explains the details of the model, starting with the core set of actors that feature in the model, before outlining the decision problems that each actor faces. We first describe the regulator's problem and then describe the different incentives that we allow governments to award them. Finally, we discuss the AI companies' problem, where we extend previous work from the literature.

\begin{figure}
	\centering
	\includegraphics[width=0.3\textwidth]{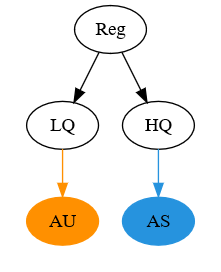}
	\caption{The Regulator's Problem in the default scenario of interest. A regulator must choose whether to invest in high-quality evaluation tools and talent, $HQ$,  or to accept a lower detection rate for unsafe practices, $LQ$. AI companies make their choice after observing the choice of the regulator. If the high-quality detection rate is high enough, then AI companies will switch from the unsafe equilibrium where they all play $\textbf{AU}$ to one where they all play $\textbf{AS}$. Via backwards induction, the regulator could reason that they are choosing over the two equilibria and will act to secure whichever equilibrium outcome is best for them.}
	\label{fig:reg-problem}
\end{figure}

\subsection{The Regulatory Market Model}

Regulatory Markets involve 3 core sets of actors:
\begin{itemize}
    \item Governments who set targets for private regulators to meet and, therefore, have oversight on what regulators test for. Governments licence private regulators to provide oversight of AI firms in their markets.
    \item Private regulators compete with each other for AI companies to choose them to provide oversight. They may compete to meet government targets. Regulators may have a wider array of powers to enforce their regulation than existing private regulators tend to, including imposing fines, requiring audits, and revoking licences.
    \item AI companies who must choose from available Private Regulators for their desired market(s). The requirements of Regulators are mandatory.
\end{itemize}

We have only two populations in the baseline model, regulators and AI companies. For simplicity, we assume that one external government is responsible for setting the incentives facing private regulators. This government entity is also assumed to have sufficient institutional power to enforce that AI companies work with at least one regulator should they wish to deploy their advanced AI systems.

We might assume that there will be many fewer regulators than AI companies, although this will depend on a number of choices. Are we considering a wide range of possible AI companies, or only a select few who have dedicated their innovative efforts to creating General Purpose AI systems \textemdash{} the scope of AI companies matters? Barriers to entry may limit the number of AI companies in especially lucrative and risky domains \parencite{bar_formerly_borkovsky_2009dynamic, askell2019role}.

On the other hand, the number of regulators may depend largely on the degree of success that a Regulatory Market proposal has in encouraging the creation of private regulators. Will developers at existing AI companies leave to create start-ups in the market for AI regulation? Will such start-ups be sustainable or avoid buyout from AI companies? Or will the market for private regulators mainly be carved up by existing institutions \parencite{clark2019regulatory, hollenbeck2020horizontal}?

\bigskip

 \begin{table}[ht]
 \small
\centering
\begin{tabular}{c p{0.4\linewidth} c}
\toprule   
Symbol & Definition & Range\\
 \midrule
$b$ & Short term value of market & $4$ \\
$B$ & Long term value of market & $> 0$ \\
$c$ & Cost of safety measures for firms & $1$ \\
$W$ & Length of time to develop transformative AI safely & $> 0$ \\
$s$ & Speed Advantage of skipping safety precautions & $> 0$ \\
$p_r$ & The risk of disaster if a firm is Unsafe. & [0, 1] \\
$p$ & $1 - \textrm{risk of disaster}$ & [0, 1]\\
$p_l$ & The chance of a low-quality regulator & 0\\
 & revealing an unsafe firm & \\
$p_h$ & The chance of a high-quality regulator revealing an unsafe firm & $1 > p_h  > p_l = 0$ \\
$\phi$ & The regulator’s impact on the speed of unsafe firms they catch & [0, 1]\\
$g$ & Government budget allocated to regulators per firm regulated & $> 0$ \\
$r_l$ & Net profit for regulator (low-quality) & $0$ \\
$r_h$ & Net profit for regulator (high-quality) & $-1$ \\
$\beta$ & Learning rate & $0.02$\\
$Z_{reg}$ & Size of Regulatory Market & $50$\\
$Z_{ai}$ & Size of AI market & $50$\\
\bottomrule
\end{tabular}
 \bigskip
\caption{Parameter Table \textemdash{} Several of the parameters are fixed because previous work on similar models has revealed that they have little influence on the results.}
\label{tab:parameters-1}
\end{table}

\subsection{The strategic interaction between AI Companies}

AI companies enter into competition with each other and are matched to a relevant regulator from the Regulatory Market. Crucially, we assume that they first observe the regulator’s choice of effort before choosing how safe to be. We assume that they can follow one of three strategies:

\begin{itemize}
    \item \textbf{AS}  \textemdash{} companies always develop AI systems safely.
    \item \textbf{AU} \textemdash{} companies never allocate effort to AI Safety.
    \item \textbf{VS} \textemdash{} companies develop their AI systems safely, but only if they observe that regulators have invested in high-quality vetting systems.
\end{itemize}

\hfill \break

 \begin{table}[ht]
\centering
\begin{tabular}{@{\extracolsep{4pt}}lcc}
\toprule   
Strategy & Always Safe (AS) & Always Unsafe (AU) \\
 \cmidrule{2-2}
 \cmidrule{3-3}
% \midrule
AS & $\frac{B}{2W} - c$ & $p_h \cdot \frac{1}{\phi + 1} \frac{B}{W} - c$ \\
AU  & $p \cdot (1-p_h) \cdot \frac{B}{W} + p_h \frac{\phi}{\phi + 1} \frac{B}{W}$  & $p (1-p^2_h) s \frac{B}{2W} + p_h^2\frac{\phi}{\phi+1} \frac{B}{2W}$ \\
\bottomrule
\end{tabular}
 \bigskip
\caption{AI Company Payoff Matrix \textemdash{} These payoffs capture the payoffs of different strategies. Notice that the conditional strategy is not included since depending on the Regulator's choice it performs identically to one of the other strategies. We use the detection rate for high-quality regulators. Against a low-quality regulator, the payoff matrix is the same, except that we replace the detection rate with that for low-quality regulators. See the main text for explanations of each symbol. Note that the short-term benefit of producing AI systems, $b$, has been omitted to ensure a more readable table. The omitted parameters do not influence our results.} 
\label{tab:payoffs-companies}
\end{table}

Table~\ref{tab:payoffs-companies} describes the average payoffs that AI companies receive when faced with another company playing a particular strategy, given the choice of the regulator. This model is heavily based on the DSAIR model from \parencite{han2020regulate}. Companies who are always safe, $\textbf{AS}$, are at a disadvantage against companies who take risks, $\textbf{AU}$, so it is usually the unsafe firm who is the first to bring the new AI capability to market, winning the big prize, $B$ (note that payoffs are averaged over the length of the competition \textemdash{} which is $W$ if firms are safe, or $\frac{W}{s}$ if the winner is unsafe). If both companies choose the same strategy, they have an equal chance of winning the big prize.

Our model builds on the DSAIR model by adding a detection rate that differs for high- and low-quality regulators, $p_h$ is the detection rate for high-quality regulators, and $p_l$ is the detection rate for low-quality regulators. We can see from Table~\ref{tab:payoffs-companies} that increasing the detection rate (as occurs for high-quality regulators) reduces the payoffs to unsafe companies, which can only encourage them to be safer. 

Once an unsafe company is caught, the regulator (or perhaps the government) will aim to enforce that the company slows down their AI development to a speed which is a fraction $\phi$ of the safe speed. This regulatory action has an uncertain impact on who is the first to bring new AI capabilities to market. A chance remains that the previously unsafe company catches up to and overtakes the safe company, $\frac{\phi}{\phi + s}$.\footnote{To present a simpler model of payoffs, we assume that if both companies are unsafe, and caught, that they are fully punished, $\phi=0$. This has no qualitative bearing on our results, but makes the equations here much easier to interpret.}

If we restrict our attention now to this subgame played by AI companies, we can see several possible equilibrium outcomes depending on the parameters of the model. If we fix the choice of the regulator, we can ignore the conditional strategy, $\textbf{VS}$, for the time being. \cref{fig:bounty-1a} provides an illustration of the equilibria selected by social learning as we vary the risk, $p_r=1 - p$ and speed advantage, $s$, parameters of the model.

The payoffs are symmetric, so there are only a few possibilities. If the risks are high enough, then $\textbf{AS}$ is the pure strategy Nash equilibrium of the game. If the risks are low enough, $\textbf{AU}$ is the only equilibrium. If the risks are somewhere in-between, then both may be equilibria. Social learning will result in players selecting the risk-dominant equilibrium in this case. For large $\frac{B}{W}$ and a detection rate, $p_h=0$,\parencite{han2020regulate} note that $\textbf{AS}$ is risk dominant when $p > \frac{1}{3s}$. \parencite{han2020regulate} also note that society prefers companies to be safe (i.e.\ the sum of AI company utilities is greatest) whenever $p > \frac{1}{s}$. These equations give rise to a 'dilemma zone', where society prefers unsafe firms to act safely (see \cref{fig:bounty-1a}). \footnote{There are also rare choices of the values for different parameters where we can have an asymmetric equilibrium where one AI company is safe, and the other is unsafe. Our methods from Evolutionary Game Theory never select such equilibria, so we will not discuss them at greater length.} 

Now, let us allow for the choice of the regulator. As we shall discuss in more detail, the regulator can choose to be of low or high-quality: their quality determines their detection rate. It is noteworthy that if the detection rate increases due to the regulator’s choice to be high-quality, that we may move from a region of the parameter space where $\textbf{AU}$ is the only (or risk dominant) equilibrium for AI companies to a region where $\textbf{AS}$ is the only (or risk dominant) equilibrium.

There are in fact 3 relevant possibilities: firms always play $\textbf{AU}$, no matter what the regulator does, firms always play $\textbf{AS}$, and firms only play $\textbf{AS}$ if facing a high-quality regulator (this is the conditional strategy, $\textbf{VS}$).

In the first scenario, high-quality regulation is not a strong deterrent. In the second scenario, a high-quality regulator is not needed. In the third scenario, the high-quality regulator acts as a strong deterrent to unsafe behaviour, which will be socially desirable if the risk of an AI disaster is high enough, i.e.\ if we are in the dilemma zone.

\subsection{The Regulator’s Problem}

Regulators move first, with their choices fully visible to AI companies before they make their own choices. Regulators must choose whether to aim to be:

\begin{itemize}
    \item high-quality ($\textbf{HQ}$): A high-quality regulator accepts larger costs, so it has a better chance of evaluating cutting-edge AI systems. They are much more likely to detect unsafe behaviour on the part of AI companies and to know the appropriate procedures that AI companies should follow to ensure their work is safe.
    \item low-quality ($\textbf{LQ}$): They do not invest in evaluating cutting-edge AI systems, so they are unlikely to detect unsafe behaviour in more advanced systems.
\end{itemize}

For the sake of simplicity, we shall assume that they have no chance of detecting unsafe behaviour. This assumption does not affect the qualitative features of our results: What matters is that the difference between the detection rates of both regulator types is sufficient in the dilemma zone to move AI companies to develop AI safely. Since we assume that $\textbf{LQ}$ regulators essentially perform no detection services, our results also show which incentives are sufficient to encourage participation in the Regulatory Market.

\cref{fig:reg-problem} illustrates how the choice to aim for high-quality may be pivotal in influencing AI companies to develop AI more safely. Unfortunately, since high-quality regulators bear a large cost of investing in better tools and talent, this cost may outweigh any revenue they can extract from AI companies in exchange for their services. Barring government intervention, regulators will choose to be low-quality, even if it is valuable from society's point of view.

Governments could offer a flat incentive, $g$, to all regulators for each company they regulate. However, if they cannot tell which type a regulator is, then this incentive would have no effect on the choice that regulators make: it will still be more profitable to be of low-quality. 

It is clear that to design more successful incentives, governments should take into account what little information may be available.

\subsection{Bounty Incentives and Vigilant Incentives}

 \begin{table}[ht]
\centering
\begin{tabular}{@{\extracolsep{4pt}}lcccc}
\toprule   
{} & \multicolumn{2}{c}{Bounty}  & \multicolumn{2}{c}{Vigilant}\\
 \cmidrule{2-3}
 \cmidrule{4-5}
 Strategy & HQ & LQ & HQ & LQ\\ 
\midrule
Firms play \textbf{AS} & $r_{h}$ & $r_{l}$  & $r_{h} + g$ & $r_{l} + g$  \\
Firms play \textbf{AU}  & $r_{h} + g p_{h}$  & $r_{l} + g p_{l}$ & $r_{h} + g {p_h}^2$  & $r_{l} + g {p_{l}}^2 $ \\
Firms play \textbf{VS}  & $r_{h}$  & $r_{l} + g p_{l}$ & $r_{h} + g$  & $r_{l} + g {p_{l}}^2$ \\
\bottomrule
\end{tabular}
 \bigskip
\caption{Regulator Payoffs under each Incentive \textemdash{} their payoffs also depend on the regulators' efforts and on the choices made by AI companies. See the text for explanations of each strategy and the relevant parameters.} 
\label{tab:regulator-payoffs-1}
\end{table}

We consider two types of incentives. First, one could pay a bounty for any unsafe firms that regulators catch, expecting that high-quality firms will be better able to detect unsafe behaviour (“Bounty Incentives”). Second, a vigilant government could rescind a prior incentive should they discover wrongdoing on the part of the companies the regulator is responsible for (“Vigilant Incentives”).\footnote{Note that this incentive design immediately implies the need for a complementary institution which engages in monitoring the behaviour of AI companies. As \textcite{clark2019regulatory} advise, Regulatory Markets are not intended as a perfect substitute for conventional monitoring institutions}.

Table~\ref{tab:regulator-payoffs-1} presents the regulator payoffs for each type of incentive, depending on what the AI companies they regulate choose to do. Higher quality regulators always perform worse in the absence of any incentive, $r_h < r_l$ (we always set $r_l=0$ for simplicity and usually set $r_h=-1$). Bounty incentives are only on offer when AI companies play $\boldsymbol{AU}$. Bounty incentives may also be achieved when a conditionally safe AI company, $\textbf{VS}$, faces a low-quality regulator, but as we set their detection rates to $0$, this does not occur in the scenarios of interest.

We can see that when companies play $\textbf{VS}$, that $\textbf{HQ}$ regulators do as poorly as they can, while $\textbf{LQ}$ regulators do as well as they can.

The logic at play here is that Bounty Incentives would signal to private regulators that the government is willing to pay for only the tools which are effective in their job. From one point of view, there appears to be a rather appealing efficiency at play. If we only pay those regulators who actually detect unsafe behaviour, then we encourage competition to be the regulator who offers the best tools.

These Bounty Incentives certainly have their appeal in the short run in a market with rampant unsafe behaviour. Regulators will be enticed by the lucrative opportunity of catching an unsafe AI company in the act. Regulators therefore have a strong incentive to develop powerful tools for detecting a specific behaviour and may even deter the unwanted behaviour. Once the behaviour has been deterred, regulators will no longer see any profit in further improving their methods, and governments will no longer have to cover the cost of those investments.

We now turn to Vigilant Incentives. Under Vigilant incentives, all regulators receive a payment $g$. However, when AI companies play $\textbf{AU}$ or when a conditionally safe company, $\textbf{VS}$, faces a low-quality regulator, those payments will be rescinded if they fail to catch an unsafe company. We can see that when companies play $\textbf{VS}$, that $\textbf{HQ}$ regulators do as well as they can, whilst $\textbf{LQ}$ regulators do as poorly as they can.

The line of argument here is that the government chooses to treat incentives as investments in a deterrent to unsafe behaviour. A deterrent must be funded, regardless of whether unsafe behaviour is currently occurring.

Regulators are happy to receive the incentive but know that if they let unsafe companies slip away undetected, the government can claim back the incentive \textemdash{} remember that in our model, we have assumed this is the only way for the government to discriminate by regulator quality. If firms are unsafe, even high-quality regulators risk losing their incentive.

\section{Methods}\label[section]{sec:methods}

We use methods from evolutionary game theory to explore what type of behaviour the different actors within our regulatory market will learn to follow \parencite{foster1990stochastic, fudenberg2006evolutionary, wallace2015stochastic}. 

 Evolutionary Game Theory has been used to study pressing issues in AI Governance, and in Climate Change \parencite{han2020regulate, lacroix2022tragedy, encarna_cc_ao2016paradigm, santos2016evolutionary}. The field has also devoted much attention to the study of the efficiency of different incentives for resolving social dilemmas \parencite{sigmund2010social, sasaki2012take, sun2021combination, han2022institutional, cimpeanu2023social}. These methods have been used in the past to study games with multiple populations, as we do here \parencite{rand2013evolution, zisis2015generosity, santos2016evolutionary, encarna_cc_ao2016paradigm}.
 
To further motivate the use of Evolutionary Game Theory,  consider that the regulators and AI companies in our model will likely engage in a period of learning about the type of behaviour they wish to emulate. Although there may not be many AI companies with large enough capital to perform at the cutting edge, there are a wide number of applications of AI systems that these companies may wish to be active in. We anticipate that companies are uncertain about the net value of any particular new technology, and that regulators face uncertainty over how difficult it is to evaluate new technologies. In the face of this uncertainty, we expect both groups to explore the strategy space and to imitate high-performers. It seems reasonable to approximate this setting as a Moran process: we have a finite number of players who may over time randomly explore different strategies or instead imitate their more successful peers.\footnote{There is a discussion to be had about whether an evolutionary model is more or less appropriate for studying the market for future AI systems than, say, classical game theory. This may be especially relevant for the increasing competition to build large language models, which is typically led by companies willing to spend large amounts on the talent, computational infrastructure, and data collection needed to develop cutting-edge systems. Large companies may be more forward-looking and rational than smaller companies, and therefore may be less likely to learn through imitation of their peers. Nevertheless, we maintain that Evolutionary Game Theory is a useful first approximation to these scenarios, especially given the close ties between these methods, the analysis of complex agent-based models, and reinforcement learning.}

Players are more likely to imitate the strategies of players who are comparatively more successful than they are, which we capture mathematically as the difference in expected payoffs, $\Pi(k)$. Note that the expected payoffs of playing a strategy is a function of the number of players choosing other strategies. The more people play a strategy different from you, the less likely you are to interact with someone using your strategy. Your payoffs may also depend on what people are doing in other populations. In our case, the actions of regulators will influence the expected payoffs of AI companies and vice versa. We can define the success, or fitness $f$, of one strategy $A$ against $B$ as follows:

\begin{equation}
    f_{A,B}(k) = \Pi_A(k) - \Pi_B(k)
\end{equation}

To keep the analysis as straightforward as possible, we also assume that the mutation rate is infinitesimally small. In the method of \textcite{fudenberg2006imitation} this assumption is made so that in the long run the evolutionary system spends all its time in one of its absorbing states. Analyses which make this assumption often find results applicable well beyond the strict limit of very small mutation (or exploration) rates \parencite{hauert2007via, sigmund2010social, rand2013evolution}.

\begin{figure}
	\centering
	\includegraphics[width=0.6\textwidth]{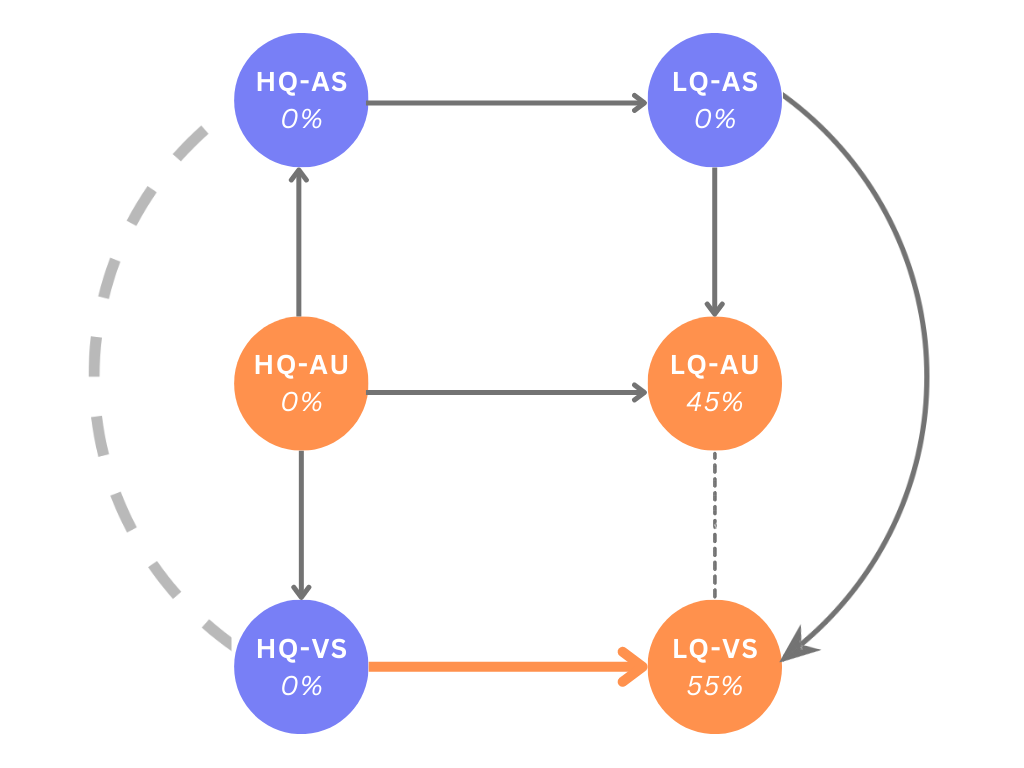}
	\caption{Bounty Incentives allow unsafe AI companies to exploit the presence of low-quality regulators. This Markov Chain diagram shows the transitions between states and their long-term frequencies. States are coloured blue if AI companies act safely and orange if AI companies act unsafely. The parameters chosen place us in the dilemma zone, $p_h=0.6$, $g=1.2$, $\phi=0.5$, $p_r=0.6$, $s=1.5$, $B/W=100$, $\beta=0.02$.}
	\label{fig:markov-chain-bi}
\end{figure}

Recall that this is a model with multiple populations, so the absorbing states are any states where all regulators follow the same behaviour and all AI companies follow the same behaviour: $\textbf{HQ-AS}$, $\textbf{HQ-AU}$, $\textbf{HQ-VS}$, $\textbf{LQ-AS}$, $\textbf{LQ-AU}$, $\textbf{LQ-VS}$. These states are visible in \cref{fig:markov-chain-bi}.

When one of these rare mutations does occur, we can calculate the likelihood that the single mutant will invade the relevant population, i.e.\ the fixation probability, using the following formula

\begin{equation}
    \rho_{A,B} = \frac{1}{1 + \sum_{j=1}^{N-1}{e^{\beta \sum_{k=1}^{j} f_{A,B}(k)}}}.
\end{equation}

In the equation above, $\beta$ refers to the imitation rate. A larger $\beta$ means players are more inclined to imitate a more successful player's strategy. Different values of $\beta$ may be appropriate in different contexts. For our figures, we choose a value for $\beta$ which implies that players are at least 90\% likely to adopt a strategy which gives payoffs one standard deviation greater than their current strategy, which seems reasonable given the high stakes involved, especially for companies. A sensitivity analysis of the figures we present in this paper suggests that our results are robust to different choices of $\beta$ (we chose values of $\beta$ which instead implied a 75\% and 95\% adoption likelihood). The value of $\beta$ can also be informed through behavioural experiments with human participants \parencite{rand2013evolution, zisis2015generosity, hoffman2015experimental}.

We can also see from \cref{fig:markov-chain-bi} that if we want to know how much time we spend in each state on average, we should care about the transitions between each of these states. Assuming that all mutations are just as likely to occur, it is straightforward to derive a transition matrix to tell us the relative likelihood with which the system is likely to move from one state to another. Due to the rare mutation limit, only one population, AI companies or regulators, will experience a mutation during a given evolutionary epoch. Therefore, we only need to consider transitions between states where one of the populations remains unchanged.

We can write the elements of the transition matrix as follows, where $S$ is the number of states:\footnote{Note that the fixation rate used for each element is the one relevant to the population undergoing the transition. If regulators changed strategy, then the fixation rate considers the success of the regulator's new strategy against the old one, rather than the success of an AI company's strategy which is unaffected by the transition.}

\begin{equation}
    P_{ij} = 
    \begin{cases}
    \frac{\rho_{ij}}{S-1} \textrm{ if } i \ne j \textrm{ and both states differ for one population only}, \\
    1 - \sum_{k = 1, k \neq i}^S{\frac{\rho_{ik}}{S-1}} \textrm{ if } i = j, \\
    0 \textrm{ if states i and j differ for more than one population.}
    \end{cases}
\end{equation}

By construction, this transition matrix is irreducible. Therefore, this transition matrix has a unique stationary distribution (see \textcite{h_aggstr_om2002finite} for a proof). This unique stationary distribution, $V$, satisfies

\begin{equation}
    (I - V) P = \vv{0}.
\end{equation}

We can find this unique stationary distribution by noting that $V$, is the normalised left eigenvector with eigenvalue $1$ of the transition matrix, $P$. We use the Grassmann-Taksar-Heyman (GTH) algorithm to compute these eigenvectors in our numerical calculation \parencite{stewart2009probability}.

The stationary distribution can be interpreted as the percentage of time that the system spends in (or around) each of these states. The results to follow interpret the stationary distribution as telling us the relative frequencies with which each type of interaction occurs between a regulator and AI companies.

\section{Results}\label[section]{sec:results}

We now turn to our analytical and numerical results. In the first two sections, we explain our key takeaways concerning Bounty and Vigilant incentives. After a brief discussion of why differences arise between our analytical and numerical results, we then consider the optimal design of a Regulatory Market when balancing the competing concerns of risk reduction and overregulation.  We then discuss how beliefs about AI risks and the size of externalities might influence the optimal design. Finally, we compare a Regulatory Market with Vigilant Incentives to direct government regulation.

\begin{figure}
	\begin{subfigure}{0.5\textwidth}
    	\centering
    	\includegraphics[width=\textwidth]{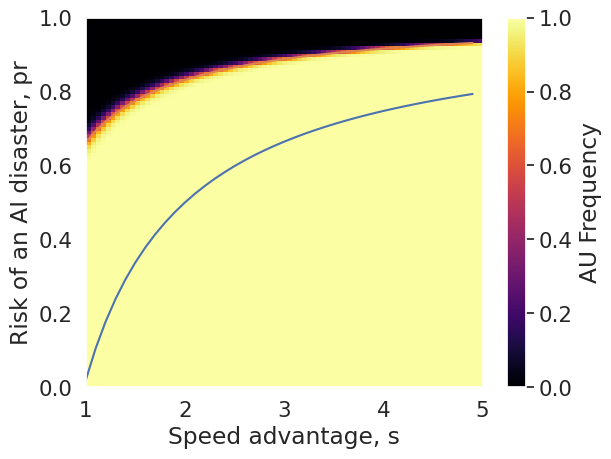}
    	\caption{AU Frequency}
    	\label{fig:bounty-1a}
	\end{subfigure}
    \hspace{1em}
	\begin{subfigure}{0.5\textwidth}
    	\centering
    	\includegraphics[width=\textwidth]{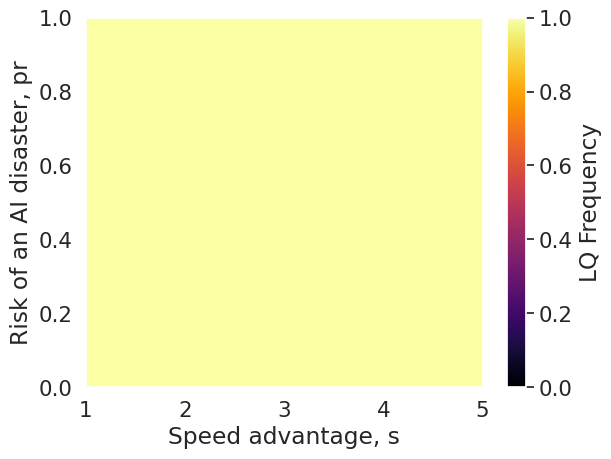}
    	\caption{LQ Frequency}
    	\label{fig:bounty-1b}
	\end{subfigure}
	\caption{Bounty Incentives have negligible impact on the behaviour of regulators and companies. (\textbf{Panel a}) The parameter space (here we show the speed advantage, $s$, and level of AI risk, $p_r$) can be split into regions where AI companies are Always Safe or Always Unsafe. AI companies choose their behaviour as they would have in the absence of any Regulatory Market. The solid lines indicate the risk dominance (top line) and socially efficient thresholds (bottom line) for the always safe strategy in the absence of a Regulatory Market. The area between them is the “dilemma zone”. (\textbf{Panel b}) No regulator invests in high-quality tools. We would therefore not see any change in welfare relative to a scenario where the government incentive and high-quality detection rate are both $0$. The model parameters take on values: $p_h=0.6$, $g=1.2$, $\phi=0.5$, $B/W=100$, $\beta=0.02$.}
	\label{fig:bounty-1}
\end{figure}

\subsection{Bounty Incentives fail to sustain a Regulatory Market}

When we first introduced Bounty Incentives, we told a plausible story of why they would be appealing to introduce. Our first result shows that this intuition was misguided.

\cref{fig:bounty-1a} suggests that catching unsafe AI companies in the act is only a dream. AI companies know to play it safe when a high-quality regulator is active, so regulators cannot benefit by improving their detection rate. \cref{fig:bounty-1b} confirms that in the long run, regulators learn that they are better off skipping the investment, and AI companies remain unsafe in the dilemma zone.

Let us consider a brief analysis of the model. The subgame-perfect Nash equilibrium of the game can be solved by backwards induction. Assume that we have model parameters such that \cref{fig:reg-problem} describes the relevant equilibria for AI companies when faced with different regulators. Therefore, each company's strategy is to play $\textbf{AU}$ when facing a $\textbf{LQ}$ regulator and to play $\textbf{AS}$ when facing a $\textbf{HQ}$ regulator. This is precisely the conditional strategy, $\textbf{VS}$.

The regulator will choose whichever option leads to an equilibrium with greater payoff. Let $I(.)$ denote a function that maps the detection rates $p_h$ or $p_l$ to the size of the incentive they expect to receive. Thus, the regulators will choose $\textbf{HQ}$ if $r_h + I(p_h | \textrm{companies play \textbf{AS}}) > r_l + I(p_l | \textrm{companies play \textbf{AU}})$. We assume regulator profits are lower for high-quality regulators, $r_h < r_l$, so we need to choose incentives which are increasing in either the detection rate or in the number of safe companies.

The Bounty incentive achieves neither of these features, $r_h < r_l + g * p_l$, and if the low-quality detection rate is positive, discourages regulator investment. In the SPNE, regulators will be of low-quality.

The Markov chain diagram, \cref{fig:markov-chain-bi}, conveys the dynamics at play in the dilemma zone of \cref{fig:bounty-1a}. Regulators who try to be $\textbf{HQ}$ will find that AI companies will either move to the conditional strategy, $\textbf{VS}$ or play $\textbf{AS}$. In either case, the regulator has no unsafe firms to catch, so it switches to $\textbf{LQ}$. If AI companies were playing $\textbf{AS}$, they would return to an unsafe strategy once the regulator gave up. Once in these states, $\textbf{LQ-AU}$ or $\textbf{LQ-VS}$, regulators and AI companies are extremely unlikely to change what they do.

We can explain these dynamics in relation to our story from earlier. As mentioned above, regulators eventually become complacent once they have deterred companies from acting unsafely. However, progress in AI often leads to new capabilities, capabilities which may require regulators to consistently improve and innovate on their approach to detecting unsafe practices. It is easy to imagine that new risks will go relatively unnoticed by regulators, similar to the situation credit rating agencies found themselves in prior to the 2007-8 financial crisis  \parencite{clark2019regulatory}. Incumbent regulators may eventually consider improving their capabilities, but as AI companies learn to behave conditionally on observing this effort. They will quickly lose interest in further investment, as they cannot find unsafe companies to collect a bounty for. These incumbent regulators will only give us a false impression of safety.

\begin{figure}
	\centering
	\includegraphics[width=0.6\textwidth]{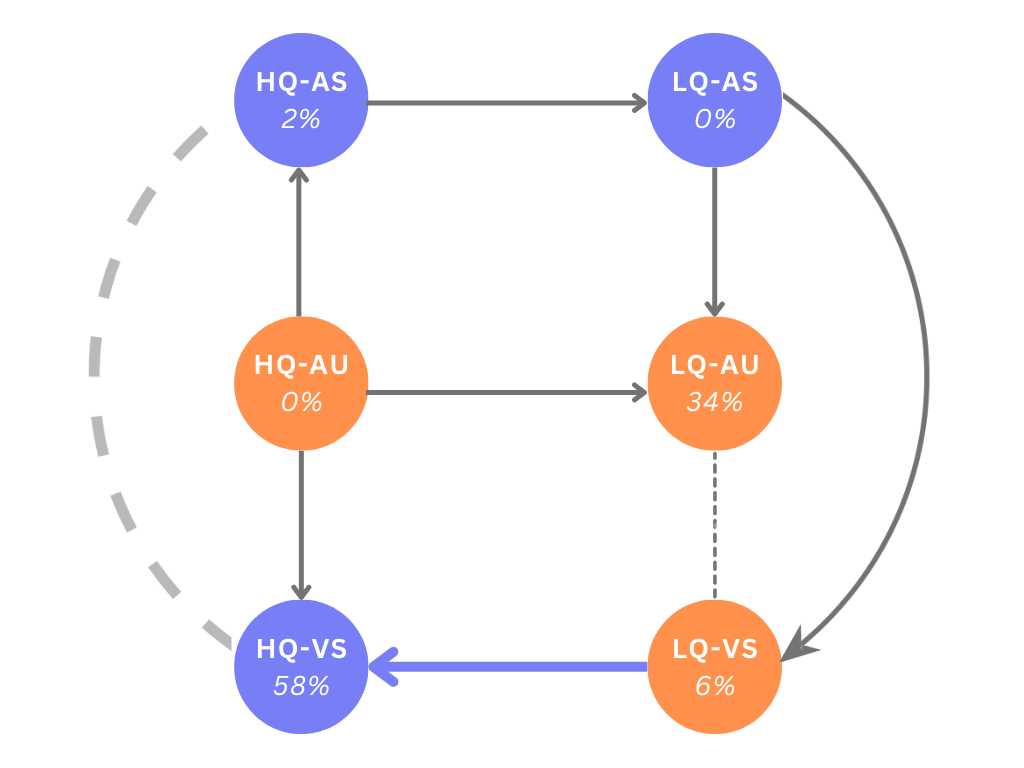}
	\caption{Vigilant Incentives encourage regulators to be high-quality innovators. This Markov Chain diagram shows the transitions between states and their long-term frequencies. States are coloured blue if AI companies act safely and orange if AI companies act unsafely. The parameters chosen place us in the dilemma zone, $p_h=0.6$, $g=1.2$, $\phi=0.5$, $p_r=0.6$, $s=1.5$, $B/W=100$, $\beta=0.02$.}
	\label{fig:markov-chain-vi}
\end{figure}

\begin{figure}
	\begin{subfigure}{0.4\textwidth}
    	\centering
    	\includegraphics[width=\textwidth]{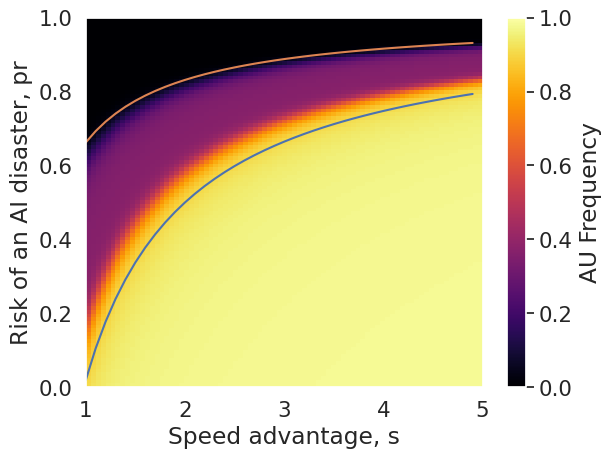}
    	\caption{AU Frequency}
    	\label{fig:vigilant-1a}
	\end{subfigure}
	\hfill
	\begin{subfigure}{0.4\textwidth}
    	\centering
    	\includegraphics[width=\textwidth]{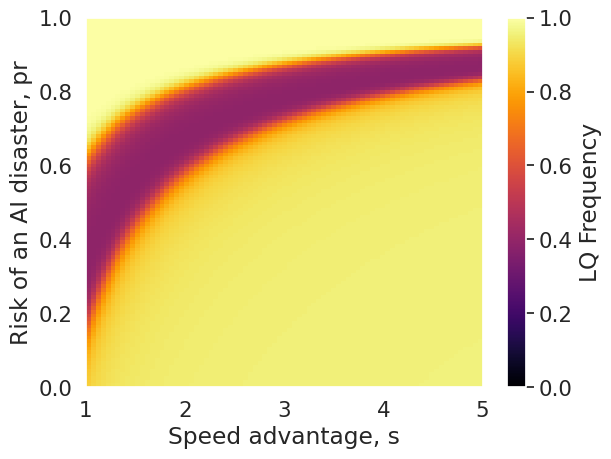}
    	\caption{LQ Frequency}
    	\label{fig:vigilant-1b}
	\end{subfigure}
	\hfill
	\begin{subfigure}{0.4\textwidth}
    	\centering
    	\includegraphics[width=\textwidth]{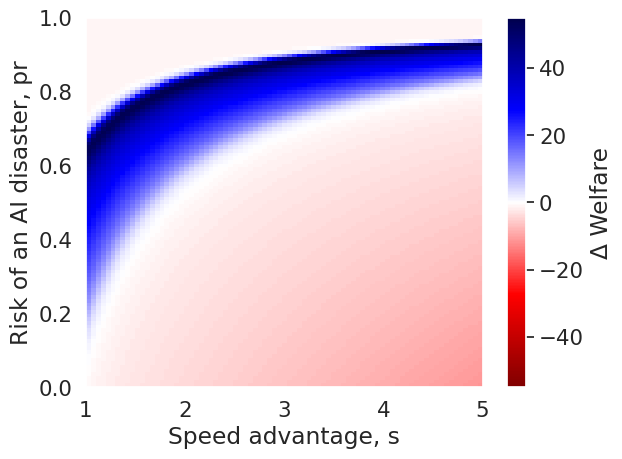}
    	\caption{$\Delta$ Welfare}
    	\label{fig:vigilant-1c}
	\end{subfigure}
	\caption{Vigilant Incentives reduce AI risk by deterring unsafe behaviour. (\textbf{a}) The parameter space (here we show the speed advantage, $s$, and level of AI risk, $p_r$) can be split into regions where AI companies are Always Safe or Always Unsafe. AI companies choose their behaviour as they would have in the absence of any Regulatory Market. The solid lines indicate the risk dominance (top line) and socially efficient thresholds (bottom line) for the always safe strategy in the absence of a Regulatory Market. The area between them is the “dilemma zone”.(\textbf{b}) Regulators often choose to be high-quality in the dilemma zone”. A small percentage of regulators remain high-quality outside of it.(\textbf{c}) The Regulatory Market improves welfare in the dilemma zone, but slightly reduces welfare through overregulation outside of it. The model parameters take on values: $p_h=0.6$, $g=1.2$, $\phi=0.5$, $\frac{B}{W}=100$, $\beta=0.02$.}
	\label{fig:vigilant-1}
\end{figure}

\subsection{Vigilant Incentives are sufficient to sustain a Regulatory Market that deters unsafe behaviour}

We now consider our “Vigilant incentive”, which satisfies the requirements of our preferred SPNE. $HQ$ is the SPNE whenever $r_h + g > r_l + g \cdot p_l^2$. This means that we need $g > \frac{r_l - r_h}{1 - p_l^2}$. If the detection rate of low-quality firms is $0$, then the government incentive only needs to be large enough to cover the profit gap between low and high-quality regulators.\footnote{Notice that if we know $r_l - r_h$, we can choose $g$ such that $p_l$ has to be sufficiently low for the Regulatory Market to be worthwhile. This means that our choice of $g$ implicitly determines which values of $p_l$ are too high to justify a Regulatory Market.}

\cref{fig:vigilant-1a} confirms that such an incentive is sufficient to see the emergence of high-quality regulators as a result of social learning. We choose $g$ to be slightly larger than the SPNE needed to encourage faster learning among regulators.

Consider \cref{fig:bounty-1} and \cref{fig:vigilant-1} for the Bounty and Vigilant incentives, respectively. Previously, the SPNE for Bounty incentives precisely matched the results of the evolutionary model, but the same cannot be said for Vigilant incentives: Though many regulators choose to invest in being high-quality in the dilemma zone, not all do. Moreover, even when regulators choose low-quality in the SPNE (bottom right of \cref{fig:vigilant-1b}) a very small proportion of regulators remain high-quality in the evolutionary model (with so few high-quality regulators, the different shades in \cref{fig:vigilant-1b} may be difficult to see on first glance). These results can be explained with reference to the evolutionary dynamics.

When we examine the Markov chain diagram, see \cref{fig:markov-chain-vi}, we can see that it is now much more common to be in state $\textbf{HQ-VS}$. Most importantly, regulators in the $\textbf{LQ-VS}$ state now have a reasonably strong incentive to switch to being $\textbf{HQ}$. Unlike in the previous Markov chain, there is still a somewhat probable path where companies can drift to playing $\textbf{AS}$. Regulators may become complacent, leading to lower investments after returning to the $LQ$ strategy, which in turn can allow the revival of unsafe behaviour. This is one reason why not all regulators end up being high-quality in the dilemma zone.

Another notable dynamic is the one-way transition from $\textbf{HQ-AU}$ to $\textbf{LQ-AU}$. The government incentive in our case is not high enough here to encourage regulators to be high-quality when AI companies are unsafe, helping to explain the relatively slow transition to safer states. However, if $g$ were much higher, this transition would flip, leading to a much larger fraction of regulators choosing to be high-quality even in scenarios where risks are low. Therefore, choosing a $g$ that allows for this dynamic is helpful in avoiding most overregulation but comes at the cost of accepting some proportion of unsafe firms. We discuss this trade-off more deeply in the next section.

The key takeaway here is that the design of our incentive matters. Bounty incentives fail to guard regulators against AI companies who play conditional strategies, so are unhelpful for supporting a Regulatory Market. On the other hand, Vigilant incentives reward regulators for the service they offer, discriminating against regulators who let unsafe AI companies enter the market undetected. This design ensures that when monitoring investments can deter unsafe behaviour, regulators are motivated to act on it.

\begin{figure}
	\begin{subfigure}{0.5\textwidth}
    	\centering
    	\includegraphics[width=\textwidth]{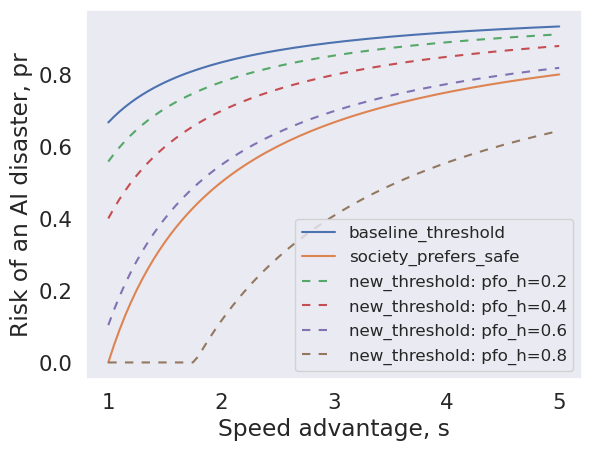}
    	\caption{}
    	\label{fig:optimal-design-1a}
	\end{subfigure}
	\hspace{1em}
	\begin{subfigure}{0.5\textwidth}
    	\centering
    	\includegraphics[width=\textwidth]{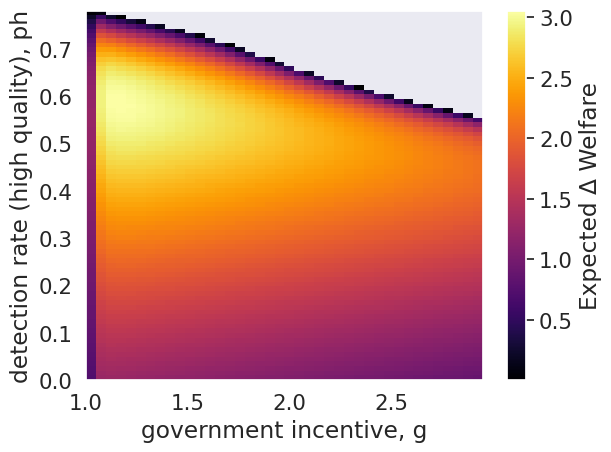}
    	\caption{}
    	\label{fig:optimal-design-1b}
	\end{subfigure}
	\caption{The details of a Regulatory Market influence welfare:
	(\textbf{a}) Risk dominance thresholds for different values of the detection rate for high-quality regulators, $g=1.2$ (other parameter values specified below).
	(\textbf{b}) Expected $\Delta$ Welfare under Vigilant Incentives for different levels of incentives, $g$, and detection rates, $p_h$. Only positive values are shown. Expected Welfare is computed uniformly over the space of $s \in [1, 5]$ and $p_r \in [0, 1]$. The model parameters take on values: $\phi=0.5$, $B/W=100$, $\beta=0.02$.}
	\label{fig:optimal-design-1}
\end{figure}

\subsection{The Optimal Design of a Regulatory Market with Vigilant Incentives}

Now that we have opted to use Vigilant Incentives to build our Regulatory Market, it is time to consider how we can optimally design this proposal to reduce risk whilst avoiding overregulation. The key lesson is that we can maintain high reductions in risk and reduce overregulation with careful tweaks to the following parameters: the detection rate that high-quality regulators achieve, $p_h$, their impact on unsafe firms they catch, $\phi$, and the government incentive, $g$. 

First, the level of detection rate, $p_h$, has a direct influence on the extent of overregulation. A well-chosen detection rate can ensure that the deterrent only has an effect when society prefers companies to be safe.

\cref{fig:optimal-design-1a} summarises this finding. Higher values of $p_h$ shift the threshold where social learning selects safe behaviour closer to the threshold where society prefers safety (the solid lines in the figure are the original and desired thresholds, as in previous figures; the dashed lines are the result of different detection rates). However, if $p_h$ is too high, then social learning will select safe behaviour even when society prefers companies to take risks. As \cref{fig:optimal-design-1a} shows, for a suitable choice of $g$, the optimal detection rate aligns the two thresholds. In this case, a detection rate of around $0.6$ seems to best align the behaviour of AI
companies with the values of society. We will later return to a discussion of when it could be feasible for policymakers to influence the detection rate.

Second, let us discuss the strength with which regulators penalise unsafe behaviour. The figures we display show results for $\phi=0.5$, indicating that regulators ensure that companies are expected to be half as slow as companies that act safely. We have also considered $\phi=1$, where companies are only slowed to match the speed of safe AI companies, and $\phi=0$ where companies are completely barred from participating in AI development. The choice of $\phi$ which is most appropriate depends on the detection rate. A high detection rate likely only needs lenient punishments to be a good deterrent to unsafe behaviour. A low detection rate requires a stronger punishment.

However, there are additional considerations which lean towards our choice of $\phi=0.5$. Lower values of $\phi$ allow a wider range of detection rates that lead to a net positive welfare effect from Regulatory Markets. These detection rates are also lower, and we should anticipate that regulators are likely to achieve lower detection rates. On the other hand, for all practical purposes, a low $\phi$ such as $\phi=0$ is implausible. History suggests that it is very difficult to bar companies from markets in which they have a strong foothold. Microsoft would be a notable example of a company who has faced legal repercussions for anti-competitive behaviour, yet to this day operates in the same markets \parencite{economides2001microsoft}. The market for AI has companies with similar levels of power and legal capability. A good compromise is likely to slow down unsafe companies by enforcing that they follow past and present safety guidance. Additional security checks and requirements could disadvantage such a company, but due to the uncertain nature of AI development, still leave them with a significant chance of catching up with safer companies. This is not to say that we should rule out the strongest punishments in all cases: If companies try something truly reckless, it may be desirable to set a new precedent and shut down their activities.

Third, \cref{fig:optimal-design-1b} suggests that the government incentive, $g$, should be large enough for high-quality regulators to perform better than others when faced with AI companies that play their conditional strategy. Otherwise, high-quality regulators would switch to being low-quality over time. In \cref{fig:optimal-design-1b}, $g > 1$ is required for this purpose: note the discontinuous jump in expected welfare past $g=1$. Additionally, regulators need to participate in the market, so they should do better than their outside option (which in our model we have assumed for simplicity to be $0$, which is the same as the net profits for low-quality regulators).

Increasing $g$ further does encourage more regulators to invest in being high-quality, which means a stronger deterrent to unsafe behaviour. However, as we can see in \cref{fig:optimal-design-1b}, a $g$ that is too high can lead to a reduction in expected welfare.
 
This reduction in expected welfare comes from overregulation, some of which is visible in the lower right corner of \cref{fig:vigilant-1c}. Higher values of $g$ ensure that regulators continue to invest in high-quality detection methods, even when the deterrent fails to deter unsafe behaviour. If the detection rate already aligns the behaviour of AI companies with society’s preferences, then these high-quality regulators will punish companies that society would prefer not to. Overregulation can be reduced by keeping $g$ low at the cost of having a higher level of unsafe behaviour in the dilemma zone.

\begin{figure}
	\begin{subfigure}{0.5\textwidth}
    	\centering
    	\includegraphics[width=\textwidth]{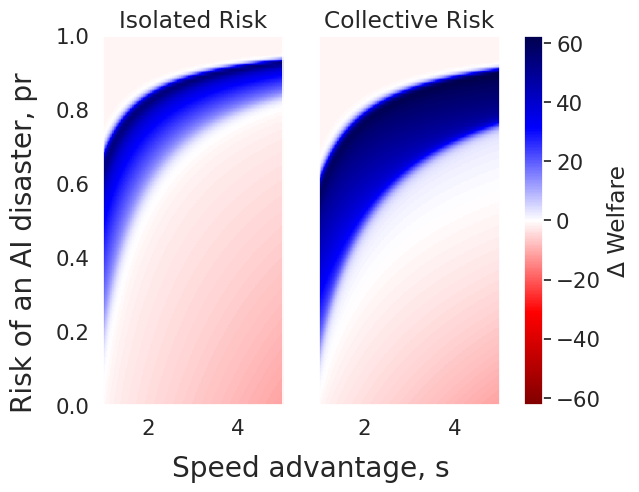}
    	\caption{No externality}
    	\label{fig:externalities-1a}
	\end{subfigure}
	\hspace{1em}
	\begin{subfigure}{0.5\textwidth}
    	\centering
    	\includegraphics[width=\textwidth]{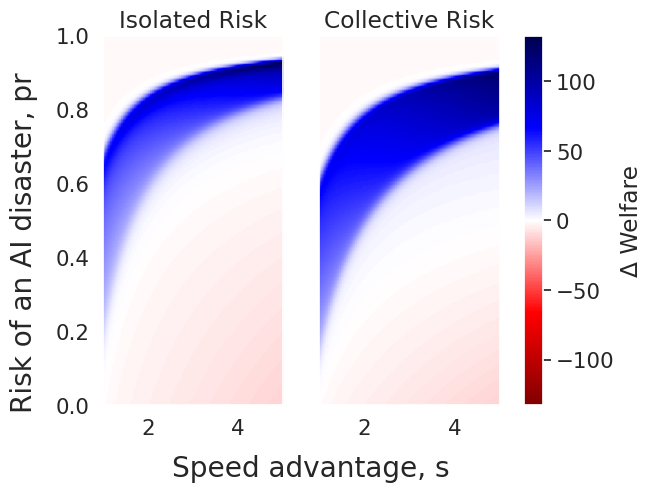}
    	\caption{Externality$=\frac{1}{5}\frac{s \cdot B}{\cdot W}$}
    	\label{fig:externalities-1b}
	\end{subfigure}
	\caption{$\Delta$ Welfare under Vigilant Incentives when we consider (\textbf{a}) adding collective risks that affect all firms, and (\textbf{b}) adding large externalities that AI companies do not expect to bear. Both ways of capturing the systemic nature of the risks presented by upcoming AI capabilities suggest that well-designed Regulatory Markets can greatly improve expected welfare. The model parameters take on values: $p_h=0.6$, $g=1.2$, $\phi=0.5$, $B/W=100$, $\beta=0.02$.}
\label{fig:externalities-1}
\end{figure}

\subsection{The need for Regulatory Markets depends on beliefs about AI risk}

Our discussion has so far considered how the design of the Regulatory Market influences its impact on social welfare. Until now, we have not explicitly discussed how different beliefs about AI risk should affect our evaluation of a Regulatory Market.

We first ask the reader to consider their beliefs about the level of risk presented by different AI capabilities, as well as the speed advantage of skipping associated safety norms. If the level of risk is high, and the speed advantage is relatively low, then we are more likely to be in the dilemma zone where society prefers unsafe AI companies to be safe. If the level of risk is low and the speed advantage is high, then it is likely that society prefers AI companies to take risks and accelerate innovation.\footnote{Another popular view is that the level of risk is increasing in the speed advantage: getting AI capabilities earlier leads to higher risk. We find that a belief that risk and speed are positively correlated places much more probability mass in the dilemma zone, giving greater justification for a Regulatory Market. However, this view has received criticism: An alternative view is that safety efforts are better targeted when AI capabilities are closer, suggesting that the level of risk may remain relatively unchanged with the speed advantage.}

Let us consider a purely illustrative example that focuses on Large Language Models, such as GPT3 and its variants. There is an argument that the risks of an AI disaster are high if these models are widely adopted and used to generate misinformation, although perhaps not above 50\%. At the same time, it might be hard to imagine seeing the pace of development we have seen so far if AI companies could not deploy LLMs until they no longer confidently generate fake information in response to a query. For illustration purposes, we might expect the mean level of risk to be normally distributed around 50\% and the speed advantage to be normally distributed around a factor of 4. As this example places most probability mass outside the dilemma zone, we should expect Regulatory Markets to mainly bring overregulation.

So far, the risks from an AI disaster have so far been assumed to be isolated to the company who enjoys the benefits of achieving breakthroughs in AI capabilities. This simplification of the model appears to be inaccurate on two accounts. First, the risks of an AI disaster may be collective in the sense that a disaster affects all companies in the AI market \textemdash{} An AI disaster may cause a government backlash and an AI winter. Or if the disaster is catastrophic, the assets or the people who make up each company may be in peril \parencite{cave2018ai, dafoe2018ai}. Second, the systemic and possibly catastrophic nature of AI risks means that externalities are very plausible. It is unlikely that AI companies will internalise the harms that misinformation or tail risks such as disruption of critical infrastructure will cause to citizens. 

\cref{fig:externalities-1} demonstrates how welfare results may change if we explicitly model the presence of externalities and the collective nature of AI disasters. If AI companies recognise the collective risk of an AI disaster, then Regulatory Markets can more directly influence behaviour. The dilemma zone is larger under collective risk because AI disasters are more likely. However, Regulatory Markets also find it easier to deter unsafe behaviour. Regulatory Markets are therefore much more likely to have a net positive welfare effect. 

If the externalities are large enough, for example $20\%$ the size of the benefits to society of having new AI capabilities as soon as possible, then Regulatory Markets are significantly better at improving welfare. Note that, as the Regulatory Market does not directly target the externality, the effect that the Regulatory Market has on behaviour does not change. 

For the most part, the results of the model are similar in pattern to those we have discussed so far. There is a wider range of scenarios where Regulatory Markets reduce risk without causing overregulation. The suggestions regarding $g$, $p_h$, and $\phi$ remain the same. However, once the externalities are large enough, overregulation ceases to be an important concern. It becomes justifiable to spend larger incentives. Higher detection rates can also be employed as the costs of overregulation become relatively small.

To summarise, Regulatory Markets become more viable when one believes the AI domain of interest has a greater risk of causing harm to society at large. If these risks are small relative to the societal benefits of having AI capabilities sooner, then a Regulatory Market would not appear to be necessary.

\begin{figure}
	\begin{subfigure}{0.5\textwidth}
    	\centering
    	\includegraphics[width=\textwidth]{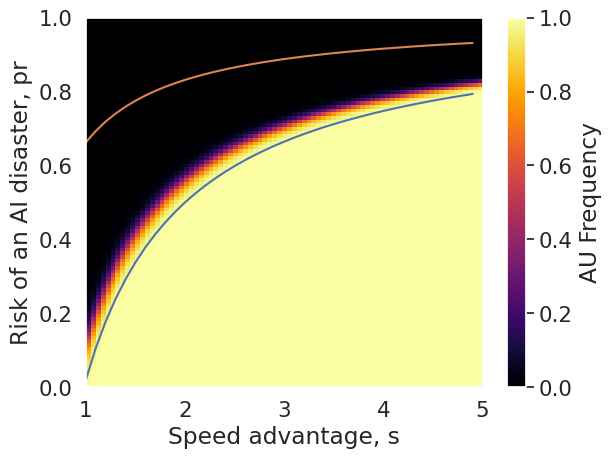}
    	\caption{AU Frequency}
    	\label{fig:gov-comparison-1a}
	\end{subfigure}
	\hfill
	\begin{subfigure}{0.5\textwidth}
    	\centering
    	\includegraphics[width=\textwidth]{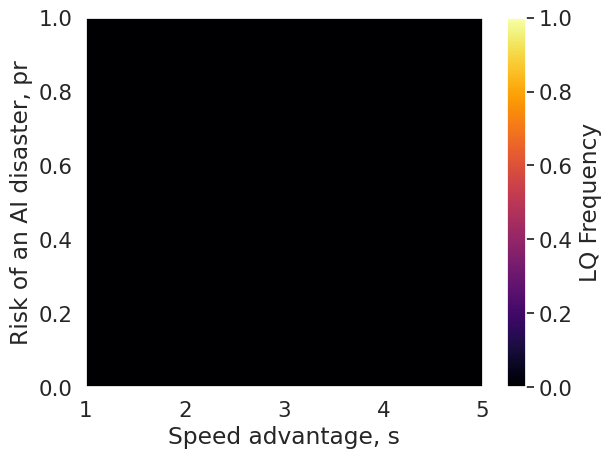}
    	\caption{LQ Frequency}
    	\label{fig:gov-comparison-1b}
	\end{subfigure}
	\hfill
	\begin{subfigure}{0.5\textwidth}
    	\centering
    	\includegraphics[width=\textwidth]{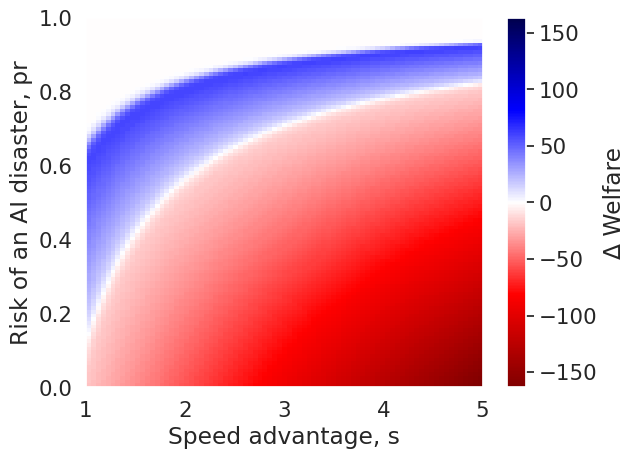}
    	\caption{$\Delta$ Welfare}
    	\label{fig:gov-comparison-1c}
	\end{subfigure}
	\caption{Direct government regulation \textemdash{} Instead of a Regulatory Market, we could allocate government spending to an institutional regulator. (\textbf{a}) Assuming they would achieve the same detection rate, they would be more effective in discouraging unsafe behaviour. The dilemma zone is completely eliminated. (\textbf{b}) However, the government always aims for high-quality. There does not exist a fallback mechanism to discourage regulation where it is not needed.(\textbf{c}) So, we see a massive loss in welfare due to overregulation outside the dilemma zone. Other parameter values are: $g=0$, $p_h=0.6$, $B/W=100$, $\phi=0.5$, $\beta=0.02$.}
	\label{fig:gov-comparison-1}
\end{figure}

\subsection{Regulatory Markets deal with uncertainty better than a Government regulator alone}

How does a Regulatory Market compare to the government enforcing regulation directly? We find results which are clearly favourable for Regulatory Markets. Though we have discussed at length the challenge of overregulation, Regulatory Markets are far better at avoiding overregulation than a government regulator.

Assume the government achieves the same optimal detection rate as we assumed our Regulatory Market does. \cref{fig:gov-comparison-1a} demonstrates that the government does better in reducing risk. In fact, since the government does not cycle between low-quality and high-quality regulators, all companies are deterred from unsafe behaviour in the dilemma zone.

However, this rigidity in the detection rate of a government regulator is also the source of overregulation, see \cref{fig:gov-comparison-1b}. Even when society prefers AI companies to take risks, the government will discover and punish these companies, slowing down innovation. Throughout this paper, we have argued that it is difficult for the government to know whether a market for AI is in the dilemma zone. Given this uncertainty, government regulation risks being excessive.

The difference between overregulation in \cref{fig:gov-comparison-1c} for the government and \cref{fig:vigilant-1c} for a Regulatory Market is substantial. Regulatory Markets do better in this case because they can fail. Regulators do not invest in better detection methods outside the dilemma zone because if they cannot deter AI companies from acting unsafely (or catch them all), the government will not pay them. Relative to direct government intervention, this failsafe means that Regulatory Markets offer a much better deal to policymakers given the uncertainty of AI development. 

We do not mean to suggest that Regulatory Markets are a replacement for government regulation. Recall that Regulatory Markets aim to meet targets set by the government in the first place. Moreover, the Vigilant Incentives we propose require that governments are knowledgeable about cutting-edge AI deployments and can independently monitor AI companies to reveal whether regulators are living up to their targets. Government monitoring is likely essential to a thriving Regulatory Market which avoids capture from the AI companies they must regulate. Ultimately, Regulatory Markets and government regulation serve as complements rather than substitutes.

\section{Discussion}\label[section]{sec:discussion}

In this paper, we have presented tentative evidence that a well-designed Regulatory Market can play a role in reducing risks from even transformative AI systems \parencite{gruetzemacher2022transformative}. Readers may also be curious about how practical considerations might inform our warnings against Bounty Incentives and our recommendations for Vigilant Incentives.

We first give additional reasons why Bounty Incentives are likely to result in the collapse of a Regulatory Market. We also suggest some obstacles to a Regulatory Market under Vigilant Incentives. These include collusion, difficulties in measuring the detection rate, and regulatory capture. We end with a brief discussion of how Regulatory Markets might be used internationally, as first suggested by the original authors of the proposal \parencite{clark2019regulatory}. The explicit modelling of these challenges and the design of tests for their presence would serve as excellent starting points for future work.

\subsection{Practical considerations for Bounty Incentives}

We have shown that Bounty Incentives tend to fail to promote investment in higher quality regulation over time. In spite of this result, we anticipate that some readers will still believe that these incentives are worth attempting, given that the government does not have to pay anything unless unsafe behaviour is caught. 

We ask our readers to contemplate the following additional reasons to avoid Bounty incentives. It could be much easier for private regulators to fake or exaggerate unsafe claims, and they might even have incentive to do so when AI companies aim to be as safe as possible. Such corruption would destroy the reputability of the Regulatory Market and would only encourage unsafe behaviour.

Another reason is that we want to avoid pitting private regulators against AI Companies as adversaries. Advocates for AI Safety are often located within AI Companies. Fostering animosity between industry and regulators only increases the difficulty of achieving consensus on the risks of future AI capabilities. We may also see other forms of antisocial punishment, such as industry or industry-aligned academics denouncing regulators \parencite{herrmann2008antisocial}. This is not to say that collusion between AI Companies and Regulators is desirable. A lack of regulatory independence could also result in ineffective regulation and may even act as a smokescreen against unacceptable behaviour \parencite{clark2019regulatory}.

Note that there may be schemes which act implicitly as Bounty Incentives. For example, a reputation system which gave high ratings of trust to regulators who detect and report the unsafe behaviour of companies could count as providing Bounty Incentives, especially if these ratings were key to the private regulators securing future lucrative work. We argue that reputation systems should aim to avoid implicit Bounty Incentives. Instead, reputation systems should focus on directly promoting truthfulness, as well as consider other insights from the literature more specific to reputation systems \parencite{barton2005who, brundage2020trustworthy, cihon2021ai}.

\subsection{Practical considerations for Vigilant Incentives}

\subsubsection{Ensuring the participation of private regulators}

Vigilant incentives may also be unappealing to regulators. It would be odd for private regulators to have a business model where failing to detect unsafe behaviour might risk the entire business (this does not have to be the case, but for simplicity we often model scenarios where high-quality regulators might not break even without government support). Stronger deterrents may lessen the risk, but care must be taken to ensure that this proposal has the ability to attract private regulators to participate in the Regulatory Market.

In our model, we simplified away the issue of participation by assuming that low-quality regulators are usually indifferent between participating in the Regulatory Market and their outside option. However, this simplification is unlikely to hold. Recent work uses Evolutionary Game Theory to show that incentivising participation is just as important as incentivising compliance for overcoming conventional social dilemmas \parencite{han2022institutional}. We should expect a similar result to hold for Regulatory Markets. 

\textcite{clark2019regulatory} argue that many of the benefits of a Regulatory Market could come from its independence from industry and the competitive pressure to find innovative ways to more cheaply evaluate cutting-edge AI systems. Both benefits seem less likely if there are high barriers to entry or if larger regulators have a motive to buy out smaller regulators. Governments can play a role in keeping the Regulatory Market competitive by incentivising new entrants, and we welcome further research on other ways governments can promote healthy competition in Regulatory Markets.\footnote{There is some measure of debate surrounding whether more concentrated markets allow for less or greater innovation, sometimes discussed as “dynamic efficiency” \parencite{demsetz1973industry, berger1998efficiency}. Companies with a high market share tend to benefit from a more inelastic demand for their products. This market power reduces the need to innovate to survive. Moreover, it appears commonplace for these companies to buy out new innovative entrants. On the other hand, companies may need the large economies of scale that a higher market share provides if they are to finance more R\&D. In addition, new start-ups might even be motivated to innovate in the hopes of a lucrative buy-out. Recent literature reveals that context matters when determining whether a pre-emptive buy-out motive overpowers high barriers to entry \parencite{hollenbeck2020horizontal}. While a stronger argument could be made in favour of market concentration for AI companies themselves, we suspect that barriers to entry and a reduced need to innovate will matter much more for private regulators, as case studies appear to suggest \parencite{clark2019regulatory}.}

\subsubsection{Reducing the cost to governments}

At first glance, Vigilant Incentives may be unappealing to governments. These incentives ask governments to at least pay each private regulator enough so that the highest-quality regulators are better off than their lower-quality analogues. As discussed, we may also need to incentivise their participation.

The literature on public goods reveals several funding mechanisms that the government can use to raise these funds from different stakeholders \parencite{tabarrok1998private, sasaki2012take, buchholz2021global}. We leave a comparison of these mechanisms in the context of a Regulatory Market to future work.

Ultimately, some groups will have to bear the cost of providing these incentives, whether they be taxpayers, AI companies, or users of AI systems. It is natural to ask if there is anything the government could do to reduce the need for these incentives in the first place.

We turn our reader's attention to an unexplored part of our setting. We assumed that in the absence of a Regulatory Market, high-quality private regulators would make a loss relative to their lower quality peers. This assumption was motivated by the larger talent and capital costs we might expect to come from investments in better detection methods for cutting-edge AI systems. We also anticipate that if we allow markets to set the price AI companies pay to private regulators, that AI companies are likely to pay more to regulators who they believe will evaluate them more favourably.

The above assumption is not guaranteed. The cost of regulatory innovation may turn out to be somewhat low for a range of AI applications. AI companies may have motives to pay more to regulators with more prestige. It might also be difficult for AI companies to win the trust of their user base if another AI company can demonstrate that their evaluation was both more relevant and more reliable. The logic here is also relevant to AI certification schemes, as discussed in \textcite{cihon2021ai}.

If the costs of evaluating AI systems are especially high, then it becomes more likely that governments can distinguish between high-quality and low-quality regulators before they perform any audits. In these cases, the government's dilemma may look very different, as they would have much more information with which to tailor their incentives.

So far, we have discussed ways the government might mitigate the cost of a Regulatory Market. However, it is worth highlighting that if the risk reduction from a Regulatory Market is high, then the costs the government faces may comparatively be very small. For this reason, we suggest that future work on Regulatory Markets consider a more thorough assessment of the costs and benefits associated with Regulatory Markets. Such work seems especially timely given that the UK Government is exploring the role of government in a similar scheme \parencite{k2023auditing}.

One more proposal that we suggest can complement Regulatory Markets is voluntary safety agreements \parencite{han2022voluntary}. Companies voluntarily make agreements to adhere to safety norms, expecting that those who violate the agreement will be punished, either by other companies or by an institution. \textcite{han2022voluntary} found in their model that voluntary safety agreements can increase safety compliance without risking overregulation. These agreements are useful because Regulatory Market incentives may need to be kept low to mitigate overregulation. Using these policies together can help eliminate the dilemma zone while keeping the costs of Regulatory Markets low.

We could add voluntary safety agreements to a Regulatory Market as follows. Besides the targets that governments would set, private regulators could enforce voluntary safety agreements that companies agree to. The increased detection capabilities of Regulatory Markets make these agreements much more credible than otherwise, since defectors from the agreement are much more likely to be caught. Keep in mind that voluntary safety agreements are ignorant of externalities; governments can set targets which take externalities into account. When in the presence of externalities, voluntary safety agreements cannot serve as a replacement for government targets that affect all companies.

\subsection{How can we discover and manipulate the detection rate?}

Unfortunately, it is not clear a priori what kind of detection rate we should expect to arise in a Regulatory Market, nor is it clear whether we can shape it.

This first challenge is empirical \textemdash{} can we know that the Regulatory Market is likely to achieve a Goldilocks detection rate that is neither too low nor too high? This issue may not be so terrible if the government is paying close attention to data on the performance of regulators. It seems plausible that the government or another independent observer could infer the detection rate of high-quality regulators. They could for example estimate the reliability of current day audits of cutting-edge AI technologies across a range of related sectors. However, a relevant objection remains: will we learn that a Regulatory Market is a good idea with enough time remaining to course-correct if necessary?

Related to this new criticism is the issue of how to influence the detection rate. A failure to detect malpractice may be the result of a lack of time or staff to perform quality checks on any audits. It could also be the result of a failure to anticipate emerging safety concerns in the latest models. If the detection rate is way too low, it's perhaps unlikely for further incentives to solve the problem \textemdash{} it may just be too difficult to expect a higher detection rate. If the detection rate is too high, governments could advise that auditors perform audits with a lower probability. However, while probabilistic spot checks seem appropriate in airline security, it will not always be appropriate for AI regulators to forgo an audit.  Alternatively, we could encourage regulators to be more forgiving. Note that for such a scheme to remain ethical, the discovered malpractice would still have to be amended. This would mean a result more in line with punishing unsafe firms less harshly. Rather than reducing the detection rate, we make larger detection rates more useful.

One takeaway is that it is useful to keep track of the detection rate of the regulators in the Regulatory Market. Besides the reasons outlined above, it is necessary to use proxy measures to gain a better idea of whether the regulators are fulfilling the government's targets. It also seems sensible to encourage as high a detection rate as possible (since it seems difficult in practise to achieve high detection rates for flaws in novel technologies). If detection rates appear to be too high, the government could recommend lighter restrictions on the companies who defect.

\subsection{Additional challenges that face a Regulatory Market}

\subsubsection{Large AI companies operate in multiple markets}

The evidence suggests that large AI companies will operate in multiple markets. Large industry-housed labs account for the vast majority of private investment into new AI capabilities \parencite{zhang2022ai}. This research stretches across multiple sectors of the economy, whether it is in improving visual effects or towards software for better robotic assistants. It is also increasingly clear that new AI capabilities allow the development of “general purpose AI systems" which we can expect on their own to be influential in multiple markets \parencite{gutierrez2022proposal}.

The challenge presented by AI companies operating in multiple markets is that it may be difficult to avoid at least one such market becoming underregulated. With so many possible applications of AI systems, it may be difficult for governments to be aware of the weakest links in their response to the risks presented by different technologies. The somewhat decentralised nature of Regulatory Markets holds promise in addressing these gaps in government monitoring, but it is not as clear from our discussion so far how governments can best target their incentives in the context of many markets.

In future work, we will integrate methods from network science to address this gap in our understanding of incentives for a Regulatory Market \parencite{choi2020large, galeotti2020targeting}. \textcite{cimpeanu2022artificial} have already studied the competitive dynamics of AI research on heterogeneous networks. Elsewhere, \textcite{cimpeanu2023social} have also studied how to target incentives to foster fairness on heterogeneous networks. However, to better represent the many markets that AI companies find themselves in, as well as to identify weak links in terms of regulations, we may need to turn to a multilayer network representation \parencite{boccaletti2014structure, walsh2019games}.

The inclusion of single and multilayer networks into the model would not only allow greater realism, it also allows for the integration of heterogeneous sources of data about the relative risks and operations of AI companies in different markets. We could use such data to inform how policymakers should allocate time and resources towards each Regulatory Market. We also open up Regulatory Markets to be tested on whether they live up to the predictions of the model \textemdash{} a failure to do so can inform governments of how they might change course. Of course, not all sources of data will be consistent. In many cases of interest, data will be missing: not all countries will have the capacity to monitor the AI landscape and not all companies will wish to be public about their research plans. In these cases, we plan to use machine learning techniques to infer the distributions relevant to the more complex model we have alluded to above.

\subsubsection{Regulatory Capture}

A big reason why regulation can fail is due to regulatory capture. Collusion between regulators and the companies they regulate may be fairly common, as often the people qualified to work in the industry have the qualifications and network needed to take up a role in the regulator. This can lead to group-think about what the risks are, in ways which might ultimately be self-serving.

It also leaves open the possibility that companies can find ways to reward regulators for approving their AI systems. If the prize is large from being the first mover in markets for future AI capabilities, then these rewards may overpower any incentives the government might offer. Regulation in this form may be worse than no regulation at all, as it may act as a smokescreen which discourages decision makers from taking pivotal action when it might be needed most.

Regulatory capture is a difficult challenge, and we do not claim to present a solution here. Nor do we explicitly model this failure mode. Nevertheless, we can make a case that Regulatory Markets, relative to a public or hybrid regulator, are more likely to avoid this capture. A thriving Regulatory Market would have many private regulators with a diverse set of overlapping responsibilities. It seems that it would be more difficult for even powerful companies to collude with most of the regulators they might work with. Additionally, the presence of an authority that complements the monitoring activities of private regulators may increase the difficulty of colluding undetected. 

In short, the careful design of incentives can encourage a Regulatory Market that operates independent of the industry, without harming relationships between AI Safety advocates and labs developing AI Capabilities.

Future work could explore a similar model to ours with an explicit collusion component \parencite{lee2019social, liu2021evolutionary}. We think it would be even more valuable for researchers to design formal tests for the presence of collusion in the market for AI in practise. Such tests should learn from models which have yielded evidence put towards antitrust cases in the past \parencite{besanko2020sacrifice}.

\subsubsection{International Coordination}

Lastly, we hope to encourage our readers to consider our discussion in an international context. \textcite{clark2019regulatory} convey that Regulatory Markets hold the most promise if they can mimic similar standard setting organisations in achieving international coordination on AI standards. There are several challenges to this endeavour.

First, as the targets that governments set for regulators are outcome based, they almost assuredly imply that AI companies will need to design their AI systems with both ethical and technical standards in mind. \textcite{von_ingersleben_seip2023competition} find that so far only technical standards for AI have seen successful adoption. \textcite{von_ingersleben_seip2023competition} attribute the failure to see international agreement on ethical standards to large differences in values between countries over these standards.\footnote{\textcite{von_ingersleben_seip2023competition} also attribute these failures to the non-excludability and non-rivalry of ethical standards \textemdash{} the properties of a public good. However, their data appears instead to support the idea that countries are after different ethical standards, rather than choosing to free-ride on producing such standards.} Barring clever framings of these difficult bargaining problems \parencite{jackson2018efficiency}, these conditions are unlikely to change.

Nevertheless, Regulatory Markets may still be successful given that each country is free to set their own targets that their regulators must adhere to. However, if Regulatory Markets in different nations ask different requirements of AI companies, then a crucial assumption of our model is broken: the model considers that all firms are equally affected by high-quality regulators. Moreover, some governments may not adopt Regulatory Markets at all, especially if they have different perceptions over the risks presented by future AI capabilities. Predictably, large economies are unlikely to enforce regulatory commitments if it puts them at a disadvantage to their economic rivals, or if it lessens their lead in a strategic domain. This narrative paints a rather bleak outlook for regulation of AI, including the Regulatory Markets proposal we discuss here.

On the other hand, we have shown that Regulatory Markets could be a useful tool for deterring unsafe behaviour, one which provides more flexibility to responding to changing national or international contexts. Just as with other measures, Regulatory Markets could serve as a commitment device in international relations, allowing governments to commit to a safer market for AI, assuming that everyone else is willing to see through similar commitments \parencite{putnam1988diplomacy, o_keefe2020windfall}. Crucially, the success of such commitments will depend on whether there is a shared perception of the risks from transformative AI \parencite{jervis1978cooperation, askell2019role}.

% \section{Conclusion}

% We used an evolutionary game theory model to explore the role governments can play in building a Regulatory Market for AI systems that deters reckless behaviour.

% We warn that it is alarmingly easy to stumble on incentives which would prevent Regulatory Markets from achieving this goal. These “Bounty Incentives” only reward private regulators for catching unsafe behaviour. We argue that AI companies will likely learn to tailor their behaviour to how much effort regulators invest, discouraging regulators from innovating. We instead recommend that governments always reward regulators, except when they find that those regulators failed to detect unsafe behaviour they should have. These “Vigilant Incentives” could encourage private regulators to find innovative ways to evaluate cutting-edge AI systems. 

% We also visualised how a Regulatory Market achieves a comparable reduction in risk to an institutional regulator whilst minimising overregulation. We conclude that Regulatory Markets with well-designed incentives can complement institutional regulators.

\vskip 0.2in
\printbibliography

@article{amodei2016concrete,
  title = {Concrete {{Problems}} in {{AI Safety}}},
  author = {Amodei, Dario and Olah, Chris and Steinhardt, Jacob and Christiano, Paul and Schulman, John and others},
  year = 2016,
  month = jul,
  journal = {arXiv},
  doi = {10.48550/arXiv.1606.06565},
  eprint = {1606.06565},
  eprinttype = {arxiv},
  primaryclass = {cs},
  archiveprefix = {arXiv}
}

@article{armstrong2016racing,
  title = {Racing to the Precipice: {{A}} Model of Artificial Intelligence Development},
  shorttitle = {Racing to the Precipice},
  author = {Armstrong, Stuart and Bostrom, Nick and Shulman, Carl},
  year = 2016,
  month = may,
  journal = {AI \& SOCIETY},
  volume = 31,
  number = 2,
  pages = {201--206},
  doi = {10.1007/s00146-015-0590-y},
  issn = {1435-5655},
  langid = {english}
}

@article{askell2019role,
  title = {The {{Role}} of {{Cooperation}} in {{Responsible AI Development}}},
  author = {Askell, Amanda and Brundage, Miles and Hadfield, Gillian},
  year = 2019,
  month = jul,
  journal = {arXiv},
  doi = {10.48550/arXiv.1907.04534},
  eprint = {1907.04534},
  eprinttype = {arxiv},
  primaryclass = {cs},
  archiveprefix = {arXiv}
}

@techreport{bar_formerly_borkovsky_2009dynamic,
  title = {A {{Dynamic Quality Ladder Model}} with {{Entry}} and {{Exit}}: {{Exploring}} the {{Equilibrium Correspondence Using}} the {{Homotopy Method}}},
  shorttitle = {A {{Dynamic Quality Ladder Model}} with {{Entry}} and {{Exit}}},
  author = {Bar (formerly Borkovsky), Ron N. and Doraszelski, Ulrich and Kryukov, Yaroslav (Steve)},
  year = 2009,
  month = oct,
  address = {{Rochester, NY}},
  doi = {10.2139/ssrn.1502860},
  type = {{{SSRN Scholarly Paper}}},
  institution = {{Social Science Research Network}},
  langid = {english}
}

@article{barrett2022actionable,
  title = {Actionable {{Guidance}} for {{High-Consequence AI Risk Management}}: {{Towards Standards Addressing AI Catastrophic Risks}}},
  shorttitle = {Actionable {{Guidance}} for {{High-Consequence AI Risk Management}}},
  author = {Barrett, Anthony M. and Hendrycks, Dan and Newman, Jessica and Nonnecke, Brandie},
  year = 2022,
  month = jun,
  journal = {arXiv},
  publisher = {{arXiv}},
  doi = {10.48550/arXiv.2206.08966},
  copyright = {arXiv.org perpetual, non-exclusive license},
  langid = {english}
}

@article{barton2005who,
  title = {Who Cares about Auditor Reputation?\textasteriskcentered},
  author = {Barton, Jan},
  year = 2005,
  journal = {Contemporary Accounting Research},
  volume = 22,
  number = 3,
  pages = {549--586},
  doi = {https://doi.org/10.1506/C27U-23K8-E1VL-20R0},
  url = {https://onlinelibrary.wiley.com/doi/abs/10.1506/C27U-23K8-E1VL-20R0},
  eprint = {https://onlinelibrary.wiley.com/doi/pdf/10.1506/C27U-23K8-E1VL-20R0}
}

@article{berger1998efficiency,
  title = {The {{Efficiency Cost}} of {{Market Power}} in the {{Banking Industry}}: {{A Test}} of the ``{{Quiet Life}}'' and {{Related Hypotheses}}},
  shorttitle = {The {{Efficiency Cost}} of {{Market Power}} in the {{Banking Industry}}},
  author = {Berger, Allen N. and Hannan, Timothy H.},
  year = 1998,
  month = aug,
  journal = {The Review of Economics and Statistics},
  volume = 80,
  number = 3,
  pages = {454--465},
  doi = {10.1162/003465398557555},
  issn = {0034-6535}
}

@article{besanko2020sacrifice,
  title = {Sacrifice Tests for Predation in a Dynamic Pricing Model: {{Ordover}} and {{Willig}} (1981) and {{Cabral}} and {{Riordan}} (1997) Meet {{Ericson}} and {{Pakes}} (1995)},
  shorttitle = {Sacrifice Tests for Predation in a Dynamic Pricing Model},
  author = {Besanko, David and Doraszelski, Ulrich and Kryukov, Yaroslav},
  year = 2020,
  month = may,
  journal = {International Journal of Industrial Organization},
  volume = 70,
  pages = 102522,
  doi = {10.1016/j.ijindorg.2019.102522},
  issn = {01677187},
  langid = {english}
}

@article{brown2021algorithm,
  title = {The Algorithm Audit: {{Scoring}} the Algorithms That Score Us},
  shorttitle = {The Algorithm Audit},
  author = {Brown, Shea and Davidovic, Jovana and Hasan, Ali},
  year = 2021,
  month = jan,
  journal = {Big Data \& Society},
  publisher = {{SAGE Publications Ltd}},
  volume = 8,
  number = 1,
  pages = 2053951720983865,
  doi = {10.1177/2053951720983865},
  issn = {2053-9517},
  langid = {english}
}

@misc{brundage2018malicious,
  title = {The Malicious Use of Artificial Intelligence: Forecasting, Prevention, and Mitigation},
  author = {Brundage, Miles and Avin, Shahar and Clark, Jack and Toner, Helen and Eckersley, Peter and others},
  year = 2018,
  publisher = {arXiv},
  doi = {10.48550/ARXIV.1802.07228},
  url = {https://arxiv.org/abs/1802.07228},
  copyright = {arXiv.org perpetual, non-exclusive license}
}

@article{brundage2020trustworthy,
  title = {Toward {{Trustworthy AI Development}}: {{Mechanisms}} for {{Supporting Verifiable Claims}}},
  shorttitle = {Toward {{Trustworthy AI Development}}},
  author = {Brundage, Miles and Avin, Shahar and Wang, Jasmine and Belfield, Haydn and Krueger, Gretchen and others},
  year = 2020,
  month = apr,
  journal = {arXiv},
  doi = {10.48550/arXiv.2004.07213},
  eprint = {2004.07213},
  eprinttype = {arxiv},
  primaryclass = {cs},
  archiveprefix = {arXiv}
}

@inproceedings{burden2020exploring,
  title = {Exploring Ai Safety in Degrees: {{Generality}}, Capability and Control},
  author = {Burden, John and {Hern{\'a}ndez-Orallo}, Jos{\'e}},
  year = 2020,
  booktitle = {Proceedings of the Workshop on Artificial Intelligence Safety ({{SafeAI}} 2020) Co-Located with 34th {{AAAI}} Conference on Artificial Intelligence ({{AAAI}} 2020)},
  pages = {36--40},
  organization = {{ceur-ws. org}}
}

@inproceedings{cave2018ai,
  title = {An {{AI Race}} for {{Strategic Advantage}}: {{Rhetoric}} and {{Risks}}},
  shorttitle = {An {{AI Race}} for {{Strategic Advantage}}},
  author = {Cave, Stephen and {{\'O} h{\'E}igeartaigh}, Se{\'a}n S.},
  year = 2018,
  month = dec,
  booktitle = {Proceedings of the 2018 {{AAAI}}/{{ACM Conference}} on {{AI}}, {{Ethics}}, and {{Society}}},
  publisher = {{Association for Computing Machinery}},
  address = {{New York, NY, USA}},
  series = {{{AIES}} '18},
  pages = {36--40},
  doi = {10.1145/3278721.3278780},
  isbn = {978-1-4503-6012-8}
}

@inproceedings{cihon2020should,
  title = {Should Artificial Intelligence Governance Be Centralised? {{Design}} Lessons from History},
  author = {Cihon, Peter and Maas, Matthijs M and Kemp, Luke},
  year = 2020,
  booktitle = {Proceedings of the {{AAAI}}/{{ACM}} Conference on {{AI}}, Ethics, and Society},
  pages = {228--234}
}

@article{cihon2021ai,
  title = {{{AI Certification}}: {{Advancing Ethical Practice}} by {{Reducing Information Asymmetries}}},
  shorttitle = {{{AI Certification}}},
  author = {Cihon, Peter and Kleinaltenkamp, Moritz J. and Schuett, Jonas and Baum, Seth D.},
  year = 2021,
  month = dec,
  journal = {IEEE Transactions on Technology and Society},
  volume = 2,
  number = 4,
  pages = {200--209},
  doi = {10.1109/TTS.2021.3077595},
  issn = {2637-6415},
  eprint = {2105.10356},
  eprinttype = {arxiv},
  primaryclass = {cs},
  archiveprefix = {arXiv},
  langid = {english}
}

@article{cimpeanu2022artificial,
  title = {Artificial Intelligence Development Races in Heterogeneous Settings},
  author = {Cimpeanu, Theodor and Santos, Francisco and Pereira, Lu{\'i}s and Lenaerts, Tom and Han, The Anh},
  year = 2022,
  journal = {Scientific Reports},
  publisher = {{Springer Science and Business Media LLC}},
  volume = 12,
  number = 1,
  pages = 1723,
  doi = {10.1038/s41598-022-05729-3}
}

@article{lee2019social,
  title = {Social Evolution Leads to Persistent Corruption},
  author = {Lee, Joung-Hun and Iwasa, Yoh and Dieckmann, Ulf and Sigmund, Karl},
  year = 2019,
  month = jul,
  journal = {Proceedings of the National Academy of Sciences},
  publisher = {{Proceedings of the National Academy of Sciences}},
  volume = 116,
  number = 27,
  pages = {13276--13281},
  doi = {10.1073/pnas.1900078116},
}

@article{liu2021evolutionary,
  title = {Evolutionary {{Dynamics}} of {{Cooperation}} in a {{Corrupt Society}} with {{Anti-Corruption Control}}},
  author = {Liu, Linjie and Chen, Xiaojie},
  year = 2021,
  month = mar,
  journal = {Int. J. Bifurcation Chaos},
  publisher = {{World Scientific Publishing Co.}},
  volume = 31,
  number = {03},
  pages = 2150039,
  doi = {10.1142/S0218127421500395},
  issn = {0218-1274}
}

@misc{zhang2022ai,
  title = {The AI Index 2022 Annual Report},
  author = {Zhang, Daniel and Maslej, Nestor and Brynjolfsson, Erik and Etchemendy, John and Lyons, Terah and others},
  year = 2022,
  month = 3,
  date = {2022-03}
}

@article{cihon2021corporate,
  title = {Corporate Governance of Artificial Intelligence in the Public Interest},
  author = {Patrick J. Cihon and Jonas Schuett and Seth D. Baum},
  year = 2021,
  journal = {Inf.},
  volume = 12,
  pages = 275
}

@article{truby2022sandbox,
  title = {A {{Sandbox Approach}} to {{Regulating High-Risk Artificial Intelligence Applications}}},
  author = {Truby, Jon and Brown, Rafael Dean and Ibrahim, Imad Antoine and Parellada, Oriol Caudevilla},
  year = 2022,
  month = jun,
  journal = {European Journal of Risk Regulation},
  publisher = {{Cambridge University Press}},
  volume = 13,
  number = 2,
  pages = {270--294},
  doi = {10.1017/err.2021.52},
  issn = {1867-299X, 2190-8249},
  langid = {english},
}

@article{gutierrez2022proposal,
  title = {A Proposal for a Definition of General Purpose Artificial Intelligence Systems},
  author = {Gutierrez, Carlos and Aguirre, Anthony and Uuk, Risto and Boine, Claire and Franklin, Matija},
  year = 2022,
  month = 10,
  day = 5,
  journal = {SSRN Electronic Journal},
  publisher = {Elsevier BV},
  doi = {10.2139/ssrn.4238951},
  date = {2022-10-05}
}

@article{boccaletti2014structure,
  title = {The Structure and Dynamics of Multilayer Networks},
  author = {Boccaletti, S. and Bianconi, G. and Criado, R. and {del Genio}, C.I. and {G{\'o}mez-Garde{\~n}es}, J. and others},
  year = 2014,
  month = nov,
  journal = {Physics Reports},
  volume = 544,
  number = 1,
  pages = {1--122},
  doi = {10.1016/j.physrep.2014.07.001},
  issn = {03701573},
  langid = {english},
}

@misc{choi2020large,
  title = {Large {{Scale Experiments}} on {{Networks A New Platform}} with {{Applications}}},
  author = {Choi, Syngjoo and Goyal, Sanjeev and Moisan, Frederic},
  year = 2020,
  pages = 117,
  langid = {english}
}

@article{galeotti2020targeting,
  title = {Targeting {{Interventions}} in {{Networks}}},
  author = {Galeotti, Andrea and Golub, Benjamin and Goyal, Sanjeev},
  year = 2020,
  month = nov,
  journal = {SSRN Electronic Journal},
  doi = {10.2139/ssrn.3054353},
}

@article{putnam1988diplomacy,
  title = {Diplomacy and domestic politics: the logic of two-level games},
  author = {Putnam, Robert D.},
  year = 1988,
  journal = {International Organization},
  publisher = {Cambridge University Press},
  volume = 42,
  number = 3,
  pages = {427–460},
  doi = {10.1017/S0020818300027697}
}

@article{jervis1978cooperation,
  title = {Cooperation under the Security Dilemma},
  author = {Jervis, Robert},
  year = 1978,
  journal = {World Politics},
  publisher = {Cambridge University Press},
  volume = 30,
  number = 2,
  pages = {167–214},
  doi = {10.2307/2009958}
}

@article{buchholz2021global,
  title = {Global {{Public Goods}}: {{A Survey}}},
  shorttitle = {Global {{Public Goods}}},
  author = {Buchholz, Wolfgang and Sandler, Todd},
  year = 2021,
  month = jun,
  journal = {Journal of Economic Literature},
  volume = 59,
  number = 2,
  pages = {488--545},
  doi = {10.1257/jel.20191546},
  issn = {0022-0515},
  langid = {english},
}

@article{cimpeanu2023social,
  title = {Social diversity reduces the complexity and cost of fostering fairness},
  author = {Theodor Cimpeanu and Alessandro {Di Stefano} and Cedric Perret and The Anh Han},
  year = 2023,
  journal = {Chaos, Solitons \& Fractals},
  volume = 167,
  pages = 113051,
  doi = {https://doi.org/10.1016/j.chaos.2022.113051},
  issn = {0960-0779},
  url = {https://www.sciencedirect.com/science/article/pii/S0960077922012309}
}

@article{clark2019regulatory,
  title = {Regulatory {{Markets}} for {{AI Safety}}},
  author = {Clark, Jack and Hadfield, Gillian K.},
  year = 2019,
  month = dec,
  journal = {arXiv},
  doi = {10.48550/arXiv.2001.00078},
  eprint = {2001.00078},
  eprinttype = {arxiv},
  primaryclass = {cs, econ, q-fin},
  archiveprefix = {arXiv}
}

@techreport{dafoe2018ai,
  title = {{{AI Governance}}: {{A Research Agenda}}},
  author = {Dafoe, Allan},
  year = 2018,
  month = aug,
  address = {{Oxford, UK}},
  institution = {{The University of Oxford}},
  langid = {english}
}

@article{demsetz1973industry,
  title = {Industry {{Structure}}, {{Market Rivalry}}, and {{Public Policy}}},
  author = {Demsetz, Harold},
  year = 1973,
  month = apr,
  journal = {The Journal of Law and Economics},
  publisher = {{The University of Chicago Press}},
  volume = 16,
  number = 1,
  pages = {1--9},
  doi = {10.1086/466752},
  issn = {0022-2186}
}

@article{economides2001microsoft,
  title = {The Microsoft antitrust case},
  author = {Economides, Nicholas},
  year = 2001,
  journal = {Journal of Industry, Competition and Trade},
  publisher = {Springer},
  volume = 1,
  pages = {7--39}
}

@article{encarna_cc_ao2016paradigm,
  title = {Paradigm shifts and the interplay between state, business and civil sectors},
  author = {Encarna\c{c}\~{a}o, Sara  and Santos, Fernando P.  and Santos, Francisco C.  and Blass, Vered  and Pacheco, Jorge M.  and others},
  year = 2016,
  journal = {Royal Society Open Science},
  volume = 3,
  number = 12,
  pages = 160753,
  doi = {10.1098/rsos.160753},
  url = {https://royalsocietypublishing.org/doi/abs/10.1098/rsos.160753},
  eprint = {https://royalsocietypublishing.org/doi/pdf/10.1098/rsos.160753}
}

@article{foster1990stochastic,
  title = {Stochastic Evolutionary Game Dynamics},
  author = {Foster, Dean and Young, Peyton},
  year = 1990,
  journal = {Theoretical Population Biology},
  volume = 38,
  number = 2,
  pages = {219--232},
  doi = {10.1016/0040-5809(90)90011-J}
}

@article{fudenberg2006evolutionary,
  title = {Evolutionary Game Dynamics in Finite Populations with Strong Selection and Weak Mutation},
  author = {Fudenberg, Drew and Nowak, Martin A. and Taylor, Christine and Imhof, Lorens A.},
  year = 2006,
  month = nov,
  journal = {Theoretical Population Biology},
  volume = 70,
  number = 3,
  pages = {352--363},
  doi = {10.1016/j.tpb.2006.07.006},
  issn = {00405809},
  langid = {english}
}

@article{fudenberg2006imitation,
  title = {Imitation Processes with Small Mutations},
  author = {Fudenberg, Drew and Imhof, Lorens A.},
  year = 2006,
  month = nov,
  journal = {Journal of Economic Theory},
  volume = 131,
  number = 1,
  pages = {251--262},
  doi = {10.1016/j.jet.2005.04.006},
  issn = {0022-0531},
  langid = {english}
}

@article{gruetzemacher2022transformative,
  title = {The Transformative Potential of Artificial Intelligence},
  author = {Gruetzemacher, Ross and Whittlestone, Jess},
  year = 2022,
  journal = {Futures},
  publisher = {{Elsevier}},
  volume = 135,
  pages = 102884
}

@misc{gursoy2022system,
  title = {System {{Cards}} for {{AI-Based Decision-Making}} for {{Public Policy}}},
  author = {Gursoy, Furkan and Kakadiaris, Ioannis A.},
  year = 2022,
  month = aug,
  publisher = {{arXiv}},
  doi = {10.48550/arXiv.2203.04754},
  eprint = {2203.04754},
  eprinttype = {arxiv},
  primaryclass = {cs},
  archiveprefix = {arXiv}
}

@book{h_aggstr_om2002finite,
  title = {Finite {{Markov}} Chains and Algorithmic Applications},
  author = {H{\"a}ggstr{\"o}m, Olle and others},
  year = 2002,
  publisher = {{Cambridge University Press}},
  volume = 52
}

@article{han2020regulate,
  title = {To {{Regulate}} or {{Not}}: {{A Social Dynamics Analysis}} of an {{Idealised AI Race}}},
  shorttitle = {To {{Regulate}} or {{Not}}},
  author = {Han, The Anh and Pereira, Luis Moniz and Santos, Francisco C. and Lenaerts, Tom},
  year = 2020,
  month = nov,
  journal = {Journal of Artificial Intelligence Research},
  volume = 69,
  pages = {881--921},
  doi = {10.1613/jair.1.12225},
  issn = {1076-9757},
  copyright = {Copyright (c) 2020 Journal of Artificial Intelligence Research},
  langid = {english}
}

@article{han2021mediating,
  title = {Mediating Artificial Intelligence Developments through Negative and Positive Incentives},
  author = {Han, The Anh and Pereira, Lu{\'i}s Moniz and Lenaerts, Tom and Santos, Francisco C.},
  year = 2021,
  month = jan,
  journal = {PLOS ONE},
  publisher = {{Public Library of Science}},
  volume = 16,
  number = 1,
  doi = {10.1371/journal.pone.0244592},
  issn = {1932-6203},
  langid = {english}
}

@article{han2022institutional,
  title = {Institutional incentives for the evolution of committed cooperation: ensuring participation is as important as enhancing compliance},
  author = {Han, The Anh},
  year = 2022,
  journal = {Journal of The Royal Society Interface},
  publisher = {The Royal Society},
  volume = 19,
  number = 188,
  pages = 20220036
}

@article{han2022voluntary,
  title = {Voluntary Safety Commitments Provide an Escape from Over-Regulation in {{AI}} Development},
  author = {Han, The Anh and Lenaerts, Tom and Santos, Francisco C and Pereira, Lu{\'i}s Moniz},
  year = 2022,
  journal = {Technology in Society},
  publisher = {{Elsevier}},
  volume = 68,
  pages = 101843
}

@article{hauert2007via,
  title = {Via freedom to coercion: the emergence of costly punishment},
  author = {Hauert, Christoph and Traulsen, Arne and Brandt, Hannelore and Nowak, Martin A and Sigmund, Karl},
  year = 2007,
  journal = {science},
  publisher = {American Association for the Advancement of Science},
  volume = 316,
  number = 5833,
  pages = {1905--1907}
}

@inproceedings{hern_andez_orallo2019surveying,
  title = {Surveying Safety-Relevant {{AI}} Characteristics},
  author = {{Hern{\'a}ndez-Orallo}, Jos{\'e} and {Mart{\'i}nez-Plumed}, Fernando and Avin, Shahar and Heigeartaigh, Sean O},
  year = 2019,
  booktitle = {Aaai Workshop on Artificial Intelligence Safety (Safeai 2019)},
  pages = {1--9},
  organization = {{CEUR Workshop Proceedings}}
}

@article{herrmann2008antisocial,
  title = {Antisocial Punishment Across Societies},
  author = {Benedikt Herrmann  and Christian Th\"{o}ni  and Simon G\"{a}chter},
  year = 2008,
  journal = {Science},
  volume = 319,
  number = 5868,
  pages = {1362--1367},
  doi = {10.1126/science.1153808},
  url = {https://www.science.org/doi/abs/10.1126/science.1153808},
  eprint = {https://www.science.org/doi/pdf/10.1126/science.1153808}
}

@article{hoffman2015experimental,
  title = {An experimental investigation of evolutionary dynamics in the Rock-Paper-Scissors game},
  author = {Hoffman, Moshe and Suetens, Sigrid and Gneezy, Uri and Nowak, Martin A},
  year = 2015,
  journal = {Scientific reports},
  publisher = {Nature Publishing Group UK London},
  volume = 5,
  number = 1,
  pages = 8817
}

@article{hollenbeck2020horizontal,
  title = {Horizontal Mergers and Innovation in Concentrated Industries},
  author = {Hollenbeck, Brett},
  year = 2020,
  journal = {Quantitative Marketing and Economics},
  publisher = {{Springer}},
  volume = 18,
  number = 1,
  pages = {1--37}
}

@techreport{jackson2018efficiency,
  title = {The {{Efficiency}} of {{Negotiations}} with {{Uncertainty}} and {{Multi-Dimensional Deals}}},
  author = {Jackson, Matthew O. and Sonnenschein, Hugo and Xing, Yiqing and Tombazos, Christis and {Al-Ubaydli}, Omar},
  year = 2018,
  month = apr,
  address = {{Rochester, NY}},
  number = {ID 3153853},
  doi = {10.2139/ssrn.3153853},
  type = {{{SSRN Scholarly Paper}}},
  institution = {{Social Science Research Network}}
}

@misc{k2023auditing,
  title = {Auditing Algorithms: {{The}} Existing Landscape, Role of Regulators and Future Outlook},
  shorttitle = {Auditing Algorithms},
  author = {{GOV.UK}},
  year = 2023,
  note = {Retrieved February 2023 from https://www.gov.uk/government/publications/findings-from-the-drcf-algorithmic-processing-workstream-spring-2022/auditing-algorithms-the-existing-landscape-role-of-regulators-and-future-outlook},
  langid = {english}
}

@misc{krakovna2020specification,
  title = {Specification Gaming: {{The}} Flip Side of {{AI}} Ingenuity},
  shorttitle = {Specification Gaming},
  author = {Krakovna, Victoria and Uesato, Jonathan and Mikulik, Vladimir and Rahtz, Matthew and Everitt, Tom and others},
  year = 2020,
  month = apr,
  note = {Retrieved February 2023 from https://deepmind.com/blog/article/Specification-gaming-the-flip-side-of-AI-ingenuity},
  langid = {english}
}

@article{lacroix2022tragedy,
  title = {The Tragedy of the {{AI}} Commons},
  author = {LaCroix, Travis and Mohseni, Aydin},
  year = 2022,
  journal = {Synthese},
  publisher = {{Springer}},
  volume = 200,
  number = 4,
  pages = 289
}

@article{leike2017ai,
  title = {{{AI Safety Gridworlds}}},
  author = {Leike, Jan and Martic, Miljan and Krakovna, Victoria and Ortega, Pedro A. and Everitt, Tom and others},
  year = 2017,
  month = nov,
  journal = {arXiv},
  publisher = {{arXiv}},
  doi = {10.48550/ARXIV.1711.09883},
  copyright = {arXiv.org perpetual, non-exclusive license},
  eprint = {1711.09883},
  eprinttype = {arxiv},
  primaryclass = {cs},
  archiveprefix = {arXiv}
}

@inproceedings{mitchell2019model,
  title = {Model {{Cards}} for {{Model Reporting}}},
  author = {Mitchell, Margaret and Wu, Simone and Zaldivar, Andrew and Barnes, Parker and Vasserman, Lucy and others},
  year = 2019,
  month = jan,
  booktitle = {Proceedings of the {{Conference}} on {{Fairness}}, {{Accountability}}, and {{Transparency}}},
  pages = {220--229},
  doi = {10.1145/3287560.3287596},
  eprint = {1810.03993},
  eprinttype = {arxiv},
  primaryclass = {cs},
  archiveprefix = {arXiv}
}

@article{naud_e2020race,
  title = {The Race for an Artificial General Intelligence: {{Implications}} for Public Policy},
  shorttitle = {The Race for an Artificial General Intelligence},
  author = {Naud{\'e}, Wim and Dimitri, Nicola},
  year = 2020,
  month = jun,
  journal = {AI \& SOCIETY},
  volume = 35,
  number = 2,
  pages = {367--379},
  doi = {10.1007/s00146-019-00887-x},
  issn = {0951-5666, 1435-5655},
  langid = {english}
}

@inproceedings{o_keefe2020windfall,
  title = {The Windfall Clause: {{Distributing}} the Benefits of {{AI}} for the Common Good},
  author = {O'Keefe, Cullen and Cihon, Peter and Garfinkel, Ben and Flynn, Carrick and Leung, Jade and others},
  year = 2020,
  booktitle = {Proceedings of the {{AAAI}}/{{ACM}} Conference on {{AI}}, Ethics, and Society},
  publisher = {{Association for Computing Machinery}},
  address = {{New York, NY, USA}},
  series = {{{AIES}} '20},
  pages = {327--331},
  doi = {10.1145/3375627.3375842},
  isbn = {978-1-4503-7110-0}
}

@article{rand2013evolution,
  title = {Evolution of fairness in the one-shot anonymous Ultimatum Game},
  author = {David G. Rand  and Corina E. Tarnita  and Hisashi Ohtsuki  and Martin A. Nowak},
  year = 2013,
  journal = {Proceedings of the National Academy of Sciences},
  volume = 110,
  number = 7,
  pages = {2581--2586},
  doi = {10.1073/pnas.1214167110},
  url = {https://www.pnas.org/doi/abs/10.1073/pnas.1214167110},
  eprint = {https://www.pnas.org/doi/pdf/10.1073/pnas.1214167110}
}

@article{santos2016evolutionary,
  title = {An {{Evolutionary Game Theoretic Approach}} to {{Multi-Sector Coordination}} and {{Self-Organization}}},
  author = {Santos, Fernando P. and Encarna{\c c}{\~a}o, Sara and Santos, Francisco C. and Portugali, Juval and Pacheco, Jorge M.},
  year = 2016,
  month = apr,
  journal = {Entropy},
  publisher = {{Multidisciplinary Digital Publishing Institute}},
  volume = 18,
  number = 4,
  pages = 152,
  doi = {10.3390/e18040152},
  issn = {1099-4300},
  copyright = {http://creativecommons.org/licenses/by/3.0/},
  langid = {english}
}

@article{sasaki2012take,
  title = {The take-it-or-leave-it option allows small penalties to overcome social dilemmas},
  author = {Sasaki, Tatsuya and Br{\"a}nnstr{\"o}m, {\AA}ke and Dieckmann, Ulf and Sigmund, Karl},
  year = 2012,
  journal = {Proceedings of the National Academy of Sciences},
  publisher = {National Acad Sciences},
  volume = 109,
  number = 4,
  pages = {1165--1169}
}

@article{shevlane2019offense,
  title = {The Offense-Defense Balance of Scientific Knowledge: {{Does}} Publishing {{AI}} Research Reduce Misuse?},
  author = {Shevlane, Toby and Dafoe, Allan},
  year = 2019,
  journal = {Proceedings of the AAAI/ACM Conference on AI, Ethics, and Society}
}

@misc{siegmann2022brussels,
  title = {The {{Brussels Effect}} and {{Artificial Intelligence}}},
  author = {Siegmann, Charlotte and Anderljung, Markus},
  year = 2022,
  month = oct,
  doi = {10.33774/apsa-2022-vxtsl},
  note = {Retrieved February 2023 from https://www.governance.ai/research-paper/brussels-effect-ai},
  type = {Preprint},
  langid = {english}
}

@article{sigmund2010social,
  title = {Social learning promotes institutions for governing the commons},
  author = {Sigmund, Karl and De Silva, Hannelore and Traulsen, Arne and Hauert, Christoph},
  year = 2010,
  journal = {Nature},
  publisher = {Nature Publishing Group UK London},
  volume = 466,
  number = 7308,
  pages = {861--863}
}

@book{stewart2009probability,
  title = {Probability, {{Markov Chains}}, {{Queues}}, and {{Simulation}}: {{The Mathematical Basis}} of {{Performance Modeling}}},
  shorttitle = {Probability, {{Markov Chains}}, {{Queues}}, and {{Simulation}}},
  author = {Stewart, William J.},
  year = 2009,
  publisher = {{Princeton University Press}},
  doi = {10.2307/j.ctvcm4gtc},
  isbn = {978-0-691-14062-9}
}

@article{sun2021combination,
  title = {Combination of institutional incentives for cooperative governance of risky commons},
  author = {Sun, Weiwei and Liu, Linjie and Chen, Xiaojie and Szolnoki, Attila and Vasconcelos, V\'{\i}tor V},
  year = 2021,
  month = 8,
  journal = {iScience},
  volume = 24,
  number = 8,
  pages = 102844,
  doi = {10.1016/j.isci.2021.102844},
  issn = {2589-0042},
  url = {https://europepmc.org/articles/PMC8334382}
}

@article{tabarrok1998private,
  title = {The Private Provision of Public Goods via Dominant Assurance Contracts},
  author = {Tabarrok, Alexander},
  year = 1998,
  month = sep,
  journal = {Public Choice},
  volume = 96,
  number = 3,
  pages = {345--362},
  doi = {10.1023/A:1004957109535},
  issn = {1573-7101},
  langid = {english}
}

@article{tabassi2021artificial,
  title = {Artificial {{Intelligence Risk Management Framework}} ({{AI RMF}} 1.0)},
  author = {Tabassi, Elham},
  year = 2021,
  month = jul,
  journal = {NIST},
  doi = {10.6028/NIST.AI.100-1},
  note = {Last Modified: 2023-02-13T09:12-05:00},
  langid = {english}
}

@article{vinuesa2020role,
  title = {The Role of Artificial Intelligence in Achieving the {{Sustainable Development Goals}}},
  author = {Vinuesa, Ricardo and Azizpour, Hossein and Leite, Iolanda and Balaam, Madeline and Dignum, Virginia and others},
  year = 2020,
  month = jan,
  journal = {Nature Communications},
  volume = 11,
  number = 1,
  pages = 233,
  doi = {10.1038/s41467-019-14108-y},
  issn = {2041-1723},
  pmcid = {PMC6957485},
  pmid = 31932590
}

@article{von_ingersleben_seip2023competition,
  title = {Competition and Cooperation in Artificial Intelligence Standard Setting: {{Explaining}} Emergent Patterns},
  shorttitle = {Competition and Cooperation in Artificial Intelligence Standard Setting},
  author = {{von Ingersleben-Seip}, Nora},
  year = 2023,
  journal = {Review of Policy Research},
  doi = {10.1111/ropr.12538},
  issn = {1541-1338},
  eprint = {https://onlinelibrary.wiley.com/doi/pdf/10.1111/ropr.12538},
  langid = {english}
}

@incollection{wallace2015stochastic,
  title = {Stochastic Evolutionary Game Dynamics},
  author = {Wallace, Chris and Young, H Peyton},
  year = 2015,
  booktitle = {Handbook of Game Theory with Economic Applications},
  publisher = {{Elsevier}},
  volume = 4,
  pages = {327--380}
}

@techreport{walsh2019games,
  title = {Games on {{Multi-Layer Networks}}},
  author = {Walsh, A. M.},
  year = 2019,
  month = jun,
  number = 1954,
  institution = {{Faculty of Economics, University of Cambridge}},
  langid = {english}
}

@article{whittlestone2021why,
  title = {Why and {{How Governments Should Monitor AI Development}}},
  author = {Whittlestone, Jess and Clark, Jack},
  year = 2021,
  month = aug,
  journal = {arXiv},
  doi = {10.48550/arXiv.2108.12427},
  eprint = {2108.12427},
  eprinttype = {arxiv},
  primaryclass = {cs},
  archiveprefix = {arXiv}
}

@article{worthington1982social,
  title = {The {{Social Control}} of {{Technology}}. {{By David Collingridge}}. ({{New York}}: {{St}}. {{Martin}}'s {{Press}}, 1980. {{Pp}}. i + 200. \$22.50.)},
  shorttitle = {The {{Social Control}} of {{Technology}}. {{By David Collingridge}}. ({{New York}}},
  author = {Worthington, Richard},
  year = 1982,
  month = mar,
  journal = {American Political Science Review},
  publisher = {{Cambridge University Press}},
  volume = 76,
  number = 1,
  pages = {134--135},
  doi = {10.2307/1960465},
  issn = {0003-0554, 1537-5943},
  langid = {english}
}

@article{zisis2015generosity,
  title = {Generosity motivated by acceptance-evolutionary analysis of an anticipation game},
  author = {Zisis, Ioannis and Di Guida, Sibilla and Han, The Anh and Kirchsteiger, Georg and Lenaerts, Tom},
  year = 2015,
  journal = {Scientific reports},
  publisher = {Springer},
  volume = 5,
  number = 1,
  pages = {1--11}
}

@misc{zwetsloot2019thinking,
  title = {Thinking {{About Risks From AI}}: {{Accidents}}, {{Misuse}} and {{Structure}}},
  shorttitle = {Thinking {{About Risks From AI}}},
  author = {Zwetsloot, Remco and Dafoe, Allan},
  year = 2019,
  month = feb,
  note = {Retrieved February 2023 from https://www.lawfareblog.com/thinking-about-risks-ai-accidents-misuse-and-structure},
  langid = {english}
}

\end{document}